%% file: main.tex
\documentclass[10pt,twocolumn,letterpaper]{article}

\usepackage{iccv}              

\input{preamble}
\definecolor{iccvblue}{rgb}{0.21,0.49,0.74}
\usepackage[pagebackref,breaklinks,colorlinks,allcolors=iccvblue]{hyperref}

\usepackage{multirow}
\usepackage[pagebackref,breaklinks,colorlinks,allcolors=iccvblue]{hyperref}
\usepackage{pifont}
\usepackage{makecell}
\usepackage{subcaption}

\title{\textit{Co-VisiON}: \textit{Co-Visi}bility Reas\textit{ON}ing on Sparse Image Sets of Indoor Scenes}

\author{
Chao Chen, Nobel Dang, Juexiao Zhang, Wenkai Sun, Pengfei Zheng,
Xuhang He, \\Yimeng Ye, Jiasheng Zhang, Taarun Srinivas, and Chen Feng\textsuperscript{\ding{41}}\thanks{\ding{41} Corresponding author.} \\
New York University \\
\texttt{cfeng@nyu.edu}
}

\begin{document}
\maketitle
\input{sec/0_abstract}    
\input{sec/1_intro}

\input{sec/2_related}
\input{sec/3_method}
\input{sec/4_baselines}

\input{sec/5_methods}

\input{sec/6_application}

\input{sec/7_conclusion}
{
    \small
    \bibliographystyle{ieeenat_fullname}
    \bibliography{main}
}

\input{sec/8_suppl}

\end{document}

%% file: sec/0_abstract.tex
\begin{abstract}

Humans exhibit a remarkable ability to recognize co-visibility—the 3D regions simultaneously visible in multiple images—even when these images are sparsely distributed across a complex scene. This ability is foundational to 3D vision, robotic perception, and relies not only on low-level feature matching but also on high-level spatial reasoning and cognitive integration. Yet, it remains unclear whether current vision models can replicate this human-level proficiency. In this work, we introduce the \textbf{Co-VisiON} benchmark, designed to evaluate human-inspired co-visibility reasoning across more than 1,000 sparse-view indoor scenarios. Our results show that while co-visibility is often approached as a low-level feature-matching task, it remains challenging for existing vision models under sparse conditions. Notably, a proprietary vision-language model surpasses all vision-only baselines, but all models fall significantly short of human performance. This gap underscores the limitations of current architectures and motivates the need for models that integrate spatial and semantic information in a human-like manner. Inspired by human visual cognition, we propose a novel multi-view baseline, \textbf{Covis}, which achieves top performance among pure vision models and narrows the gap to the proprietary VLM. We hope our benchmark and findings will spur further advancements in developing vision models capable of robust, cognitively inspired reasoning in challenging, sparse environments. Our dataset and source code can be found at \url{https://ai4ce.github.io/CoVISION}.
\end{abstract}

%% file: sec/1_intro.tex
\vspace{-4mm}
\section{Introduction}
\label{sec:intro}

Humans are remarkably adept at inferring scene layout from sparse multi-view observations by integrating spatial cues and relational context. For example, when browsing websites for real estate listings or hotel bookings, one is often presented with only a few spatially sparse images. Despite the limited visual data, humans can intuitively grasp the spatial context and deduce the relationships between images, effectively constructing a mental graph where images serve as nodes and co-visible regions form the connecting edges.

\begin{figure}[t]
  \centering
   \includegraphics[width=0.45\textwidth]{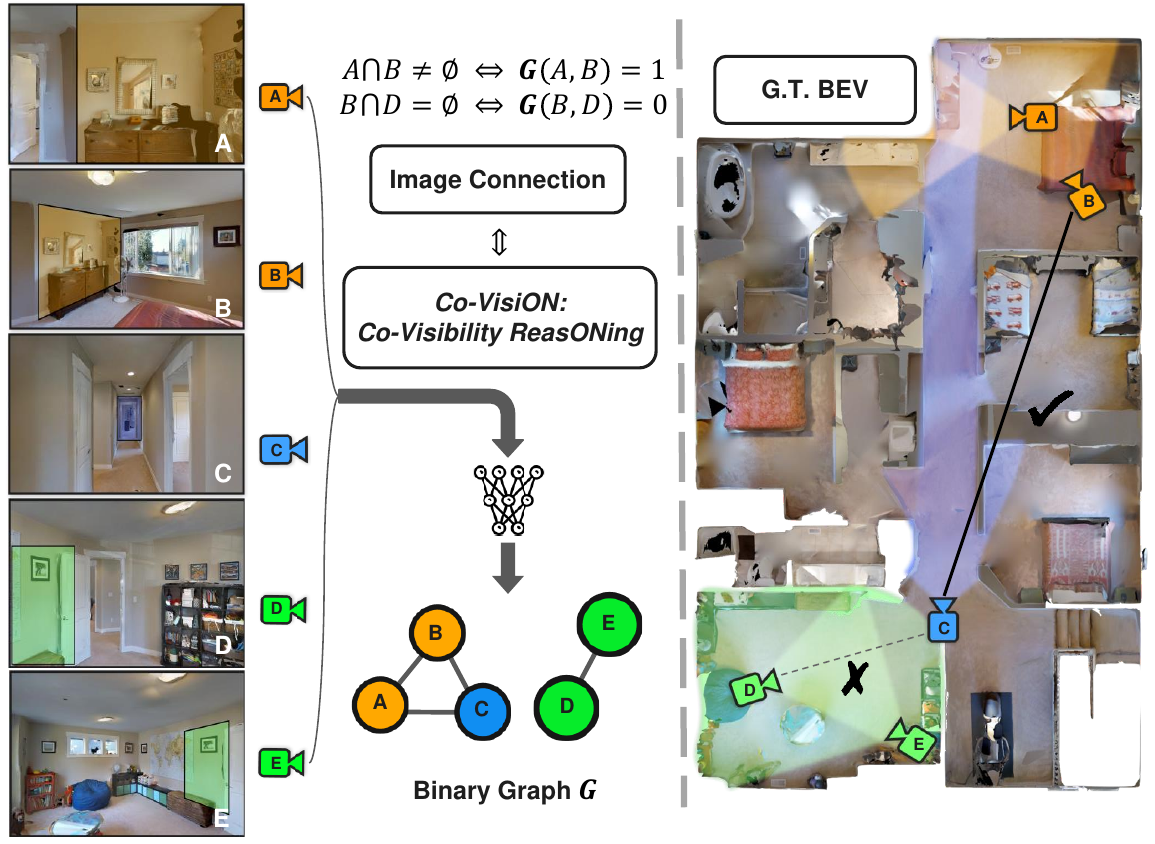}
    \vspace{-2mm}
   \caption{\textbf{Co-VisiON} overview. If two images share non-zero co-visible surface areas, they should be connected in the co-visibility graph. Note that the bird's-eye view is only for visualization and is not actually used in Co-VisiON.}
   \label{fig:teasing}
   \vspace{-6mm}
\end{figure}

We refer to this ability as \textit{co-visibility reasoning} because the key step in this process is to look for \textit{overlapping co-visible surface areas} between the images. Such areas are qualitatively illustrated in \cref{fig:teasing} with colored masks on the images on the left along with the colored field-of-views on the BEV map on the right. 
Co-visibility also plays a crucial role in 3D computer vision and robot perception, underpinning applications such as image matching, scene reconstruction, visual localization, and mapping. 

Modern vision models have shown achieved remarkable progress in these tasks and proficiency in matching co-visible images when they have dense overlaps. However, we are particularly intrigued by humans' robust co-visibility reasoning with sparsely distributed images, where shared visible information is limited or highly distorted. \textit{Can current vision models compete with human proficiency in co-visibility reasoning under sparse conditions?}

\begin{figure*}[ht!]
    \includegraphics[width=0.88\textwidth]{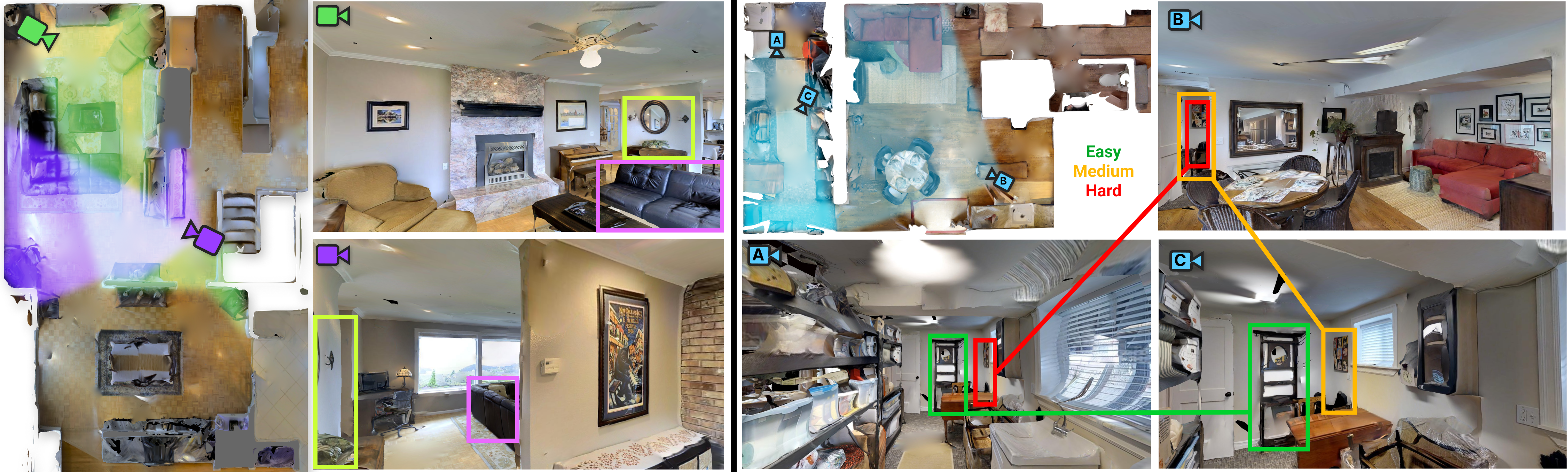}
    \vspace{-2mm}
    \centering
    \caption{\textbf{Complex Co-VisiON Examples}. This figure showcases Co-VisiON in two distinct scenarios, each featuring a selected pair, or pairs, of co-visible images that reflect explicit reasoning. On the left, weak feature correspondence and sparse angles are challenging to reason. On the right, the connection between images A and B is not obvious until the introduction of image C. These examples demonstrate the challenge in co-visibility reasoning beyond feature matching between image pairs and call for reasoning on multi-views simultaneously. The pairwise processing falls short when facing diverse viewpoints with significant angular and spatial disparities. Additional views can offer contextual information from the scene's periphery, making it easier to overcome these challenges.}
    \label{fig:examples}
    \vspace{-4mm}
\end{figure*}

In this work, we propose the \textbf{Co-VisiON} benchmark, which evaluates models on \textbf{Co-Visibility reasONing} from sparsely captured images across over 1,000 indoor scenarios. We benchmark a diverse set of relevant methods, including feature matching~\cite{lowe2004sift,fischler1981ransac,sarlin2020superglue,sun2021loftr}, contrastive learning~\cite{chen2020simple}, visual place recognition~\cite{arandjelovic2016netvlad}, 3D reconstruction~\cite{wang2023dust3r,tang2024mv}, supervised graph edge prediction~\cite{dosovitskiy2020vit, he2016resnet, simonyan2014vgg}, and large vision-language models~\cite{openai2023gpt4,yang2024qwen2}. However, our results indicate that these methods fail to capture the broader context and spatial relationships within a scene—highlighting the gap between current computational models and human-inspired spatial reasoning.

As shown in~\cref{fig:examples}, images with minimal overlap and sparse views present a significant challenge. This highlights the need for holistic understanding beyond local features, like humans.
Beyond its challenges, co-visibility reasoning is a fundamental capability for many 3D vision applications. Having a reliable co-visibility graph can significantly improve the efficiency of structure-from-motion (SfM)~\cite{ye2023ec} and simultaneous localization and mapping (SLAM)~\cite{mur2015orb} under sparse observations. Similarly, it can benefit robotics tasks such as place recognition, navigation, and pose estimation, where reasoning about spatial relationships from limited viewpoints is crucial~\cite{hutchcroft2022covispose}.

Besides the aforementioned baselines, we propose \textbf{Covis}, a supervised binary classification and mask segmentation method with either pairwise or multiple views as input. 
In our experiments, we found that the multi-view approach captures more scene information and outperforms pairwise methods when irrelevant (non-co-visible) features are excluded. Inspired by DUSt3R’s confidence scores~\cite{wang2023dust3r}, we introduce a learnable mask supervised by co-visible regions to filter out irrelevant features and improve co-visibility reasoning by focusing on the most relevant features.

In summary, our contributions are:\begin{itemize}[leftmargin=0.5cm]
\item We propose \textbf{\textit{Co-VisiON}} as a novel vision task to assess an intelligent agent's spatial reasoning ability given sparse observations, and generate a sparse indoor image dataset with dense co-visibility annotations from photorealistic indoor simulation environments.
\item We benchmark diverse baselines—including humans—and reveal the challenges and opportunities in advancing human-inspired co-visibility reasoning.

\item We demonstrate that in supervised co-visibility reasoning, the multi-view method outperforms two-view ones when assisted by co-visible mask segmentation, indicating the importance of multi-view analysis for this task.
\end{itemize}

%% file: sec/2_related.tex
\section{Related Work}
\label{sec:related}

\subsection{Related tasks}
\textbf{Visual Reasoning.} Visual reasoning tasks primarily focus on object-level relationships within a single image, such as visual question answering (e.g., GQA~\cite{hudson2018gqa}, CLEVR~\cite{johnson2017clevr}), and scene graph generation~\cite{xu2017scene, johnson2015image}. ISVQA~\cite{bansal2020isvqa} extends this to multi-image reasoning in a household setting, but it mainly targets object relationships, while Co-VisiON emphasizes image relationships grounded in the environment.

\noindent\textbf{Visual Place Recognition.} Visual place recognition (VPR) involves identifying previously seen locations and is widely studied in robotics and autonomous vehicles~\cite{lowry2015vpr, schubert2023vpr, costante2013transfer,chen2024self}. Methods like NetVLAD~\cite{arandjelovic2016netvlad} treat VPR as image retrieval with bag-of-visual-words features. The key distinction between VPR and co-visibility reasoning is in spatial requirements: VPR requires geographically close images, while Co-VisiON links images by overlapping surfaces regardless of distance or camera pose. Our experiments show that VPR-based methods are insufficient for Co-VisiON.

\noindent \textbf{Visual Localization and 3D Reconstruction.} Visual SLAM simultaneously performs image-based localization and 3D mapping from (at least) RGB cameras~\cite{bailey2006simultaneous, durrant2006simultaneous, kerl2013vslam, milford2004ratslam, milford2012seqslam, sturm12iros}, whereas Structure-from-Motion~\cite{ye2023ec, schoenberger2016sfm, pan2024global} fully reconstructs 3D scenes upto a scale and estimates camera parameters in an offline fashion. In contrast, Co-VisiON builds a co-visibility graph by linking images of a scene based on shared surface visibility, not requiring full scene reconstruction or camera pose estimation.

\noindent \textbf{Visual Correspondence.} Visual correspondence is critical for localization~\cite{mur2015orb}, image matching~\cite{sarlin2020superglue}, and scene understanding~\cite{gadre2022csr}. Traditional methods establish geometric correspondences—points~\cite{sarlin2020superglue,peng2021megloc,hutchcroft2022covispose}, lines~\cite{ye2023ec}, and planes~\cite{agarwala2022planeformers,li2024unleashing}—while recent work explores semantic ones like object relationships~\cite{gadre2022csr,gumeli2023objectmatch,elich2023learning}. Notably, Rau et al.~\cite{rau2020predicting} estimate surface overlap via image embeddings, a task related to ours but limited to image pairs and performing poorly on Co-VisiON ($20\%$ graph IoU). Overall, most correspondence methods operate on image pairs with geometric metrics, while Co-VisiON reasons over sparse sets with large pose variation and evaluates via graph IoU. 

\noindent\textbf{Related datasets}. Benchmarking Co-VisiON requires sparse image sets with ground-truth co-visibility graphs. However, most real-world datasets~\cite{dai2017scannet, yeshwanth2023scannet++,Li_2018_CVPR, rau2020predicting,silberman2012indoor}, including those in~\cite{rau2020predicting}, lack such graphs or produce noisy ones derived from coarse point clouds or depth maps. 

To address this, we construct the Co-VisiON dataset using virtual 3D scenes with textured meshes, rendered from arbitrary viewpoints with pixel-level co-visibility via OpenGL. Built on photorealistic simulation environments like Gibson~\cite{xiazamirhe2018gibsonenv} and HM3D~\cite{ramakrishnan2021habitat}, it offers sparse, well-distributed views with full scene coverage and accurate co-visible correspondences, making it well-suited for evaluating co-visibility reasoning.
\vspace{-1mm}
\subsection{Related methods}
\textbf{Feature matching.} Feature matching methods range from handcrafted descriptors like SIFT~\cite{lowe2004sift} with RANSAC~\cite{fischler1981ransac} to deep-learning approaches such as LoFTR~\cite{sun2021loftr} and SuperGlue~\cite{sarlin2020superglue}. However, we show such naive pairwise matching methods are insufficient for Co-VisiON.

\noindent \textbf{Image Representation Learning.} Image representations are commonly learned through contrastive learning, pairwise supervision, or spatially guided pretraining. Contrastive methods~\cite{oord2018infoNCE,chen2020simple,he2020momentum,wu2018unsupervised,tao2021time}, such as NetVLAD~\cite{arandjelovic2016netvlad}, aggregate local features into global descriptors. Pairwise classification approaches determine whether two images depict the same place, typically using different backbones~\cite{dosovitskiy2020vit,simonyan2014vgg,he2016resnet,Weinzaepfel_2023_ICCV}. While effective in simple settings, these methods operate only on image pairs and fail to capture broader scene context. Alligat0R~\cite{loiseau2025alligat0r}, as an example of spatially guided pretraining, and Covis both use co-visibility signals, but for different goals: pose regression vs. multi-view masking. These approaches were not designed for scene-level spatial reasoning and as our findings suggest, struggle to capture the spatial understanding.

\noindent \textbf{Reasoning with vision and language models.}
Both image matching and representation learning focus on correspondences between images from similar viewpoints or the same image with augmentations. In contrast, co-visibility reasoning requires matching images from different viewpoints, demanding spatial understanding beyond feature comparison. Since VLMs demonstrate reasoning abilities~\cite{kojima2022large, bubeck2023sparks}, we explore their potential by prompting GPT-4o~\cite{openai2023gpt4}, Gemini-2.0-Flash \cite{team2023gemini}, SpatialRGPT~\cite{NEURIPS2024_f38cb4cf}, and Qwen-72B~\cite{qwen2} for co-visibility reasoning between multiple images.

\noindent \textbf{Multi-view learning.} Multi-view learning~\cite{yu2025review} integrates information from multiple images for tasks like object-level classification~\cite{seeland2021multi,su2009learning} and 3D reconstruction~\cite{tang2024mv}. MV-DUSt3R~\cite{tang2024mv}, for example, fuses views to produce accurate 3D geometry with metric-level detail. In contrast, few works address \textit{high-level scene understanding}, which involves inferring scene topology from sparse observations. Co-VisiON tackles this underexplored setting by reasoning over multiple images to construct a co-visibility graph.

%% file: sec/3_method.tex
 \definecolor{greenpivot}{RGB}{0, 210, 0}
\vspace{-1mm}

\section{Co-VisiON Benchmark}
\label{sec:method}
\subsection{Task Definition}\label{sec:definition}
Given a sparse set of images $\mathcal{I} = \{I_1, I_2, \ldots, I_n\}$ from a scene representing different views of a house, our objective is to construct a co-visibility graph $\mathcal{G}$. In the graph, images are nodes, and they are connected by an edge only if they share co-visibility. Specifically, for each pair of images $(I_i, I_j)$, an algorithm outputs $d_{ij}$ that quantifies the degree of co-visibility between the two images.

\noindent Next, we employ thresholding to determine if $d_{ij}$ surpasses a predefined threshold $\tau$. If $d_{ij} > \tau$, we connect $I_i$ and $I_j$ with an edge in the graph $\mathcal{G}$. Covering all possible pairs of images, we construct an adjacency matrix $\mathbf{A}$, where each entry $a_{ij}$ is binarized from $d_{ij}$ by the threshold: 
\vspace{-2mm}
\begin{equation}
    a_{ij} = 1 \iff d_{ij} > \tau,
\end{equation}

indicating co-visibility between $I_i$ and $I_j$.

\noindent The resulting adjacency matrix $\mathbf{A}$ is the final representation of the co-visibility graph $\mathcal{G}$.
Therefore, by evaluating the matrix $\mathbf{A}$, a model's ability to reason co-visibility from the sparse image set can be quantitatively measured. 

\subsection{Evaluation metrics}\label{dataset_metric}
Following the notation introduced in~\cref{sec:definition}, we denote the predicted co-visibility graph and adjacency matrix by a model as $\hat{\mathcal{G}}$  and $\hat{\mathbf{A}}$ respectively. To evaluate the $\hat{\mathbf{A}}$ against the ground truth $\mathbf{A}$ of graph $\mathcal{G}$, we use two evaluation metrics; Intersection over Union (IoU) and Area Under Curve (AUC), as following:

\noindent \textbf{Graph Intersection over Union (Graph IoU):} This metric measures the similarity between $\hat{\mathcal{G}}$ and $\mathcal{G}$ by comparing the overlap of their edges~\cite{zhang2024multiview}. It is defined as the ratio of common edges to the total number of unique edges in both graphs:
$\text{Graph IoU} = \frac{|\hat{\mathbf{A}} \cap \mathbf{A}|}{|\hat{\mathbf{A}} \cup \mathbf{A}|}$.

\noindent \textbf{Area Under Curve (AUC):}
AUC is computed by averaging Graph IoU across evenly sampled thresholds in $[0, 1]$. AUC highlights fairness and captures performance robustness, where higher values indicate a more reliable co-visibility reasoning method, without relying on threshold tuning. Additional details are provided in the appendix.

\subsection{Dataset Generation}\label{sec:dataset}

To construct a structured multi-view indoor dataset with well-defined co-visibility, we employ a progressive camera placement strategy. Each image is selected to best complement scene coverage, based on its intersection over union (IoU) with previously chosen views. To ensure image diversity and graph connectivity, IoU is constrained between 5\% and 30\%. This iterative process continues until a target coverage is reached. The resulting dataset consists of a sparsely distributed image set, with a co-visibility graph encoding pairwise relationships and corresponding co-visible region information (detailed in~\cref{sec:definition}). We use Habitat-sim~\cite{szot2021habitat, habitat19iccv} to collect sparse yet representative data from Gibson~\cite{xiazamirhe2018gibsonenv} and HM3D~\cite{ramakrishnan2021habitat}, with scene-level splits to prevent data leakage. ~\cref{tab:data} presents dataset statistics, and~\cref{fig:workflow} illustrates the full dataset pipeline, with additional details provided in the supplementary material.

\noindent\textbf{Gibson}.\label{sec:Gibson} Gibson provides 3D models of real-world indoor environments, including homes, offices, and more.
We select scenes based on the availability of .glb files and required metadata. Depending on the scene size, each includes between a dozen and over a hundred images. On average, 14 images are captured per scenario across 85 distinct scenes, forming a comprehensive dataset for co-visibility graph annotation. We use an 80/20 split for training and testing.

\noindent\textbf{HM3D}.
We generate the second half of the Co-VisiON dataset from HM3D. Unlike Gibson, HM3D lacks floor height metadata. To recover vertical structure, we infer floor heights using navigable regions. The dataset is split 90\% for training and 10\% for testing.

\begin{table}[t]
\caption{\textbf{Statistics of Co-VisiON Dataset.} \textit{Scenarios} is defined as structured collections of images with associated camera poses within a given scene.}
\vspace{-3mm}
\centering
\setlength{\tabcolsep}{15pt} 
\renewcommand{\arraystretch}{0.90}  
\setlength{\extrarowheight}{0pt}    
\small                              
\begin{tabular}{lcc}
\toprule
 & \textbf{\textit{Gibson}} & \textbf{\textit{HM3D}} \\
\midrule
\textbf{1-floor scenes} & 50 & 257 \\
\textbf{2-floor scenes} & 27 & 309 \\
\textbf{3-floor scenes} & 7 & 161 \\
\textbf{$\geq$4-floor scenes} & 1 & 28 \\
\textbf{Scenarios / scene} & 5 & 1 \\
\textbf{Total scenes} & 85 & 755 \\
\textbf{Total scenarios} & 425 & 755 \\
\textbf{Total images} & 5,954 & 16,245 \\
\textbf{Total image pairs} & 33,849 & 210,008 \\
\bottomrule
\end{tabular}
\vspace{-5mm}
\label{tab:data}
\end{table}

\noindent \textbf{Human Annotations.} We collect human annotations in the Co-VisiON dataset as a benchmark baseline. For each Gibson scene, several experienced annotators generate a co-visibility graph, which is then compared to the ground truth to compute graph IoU. This provides an estimate of human-level performance on the Co-VisiON task.

%% file: sec/4_baselines.tex
\section{Experiments on Co-VisiON}
\label{sec:experiment}

\begin{figure*}[ht!]
    \centering
    \includegraphics[width=0.94\textwidth]{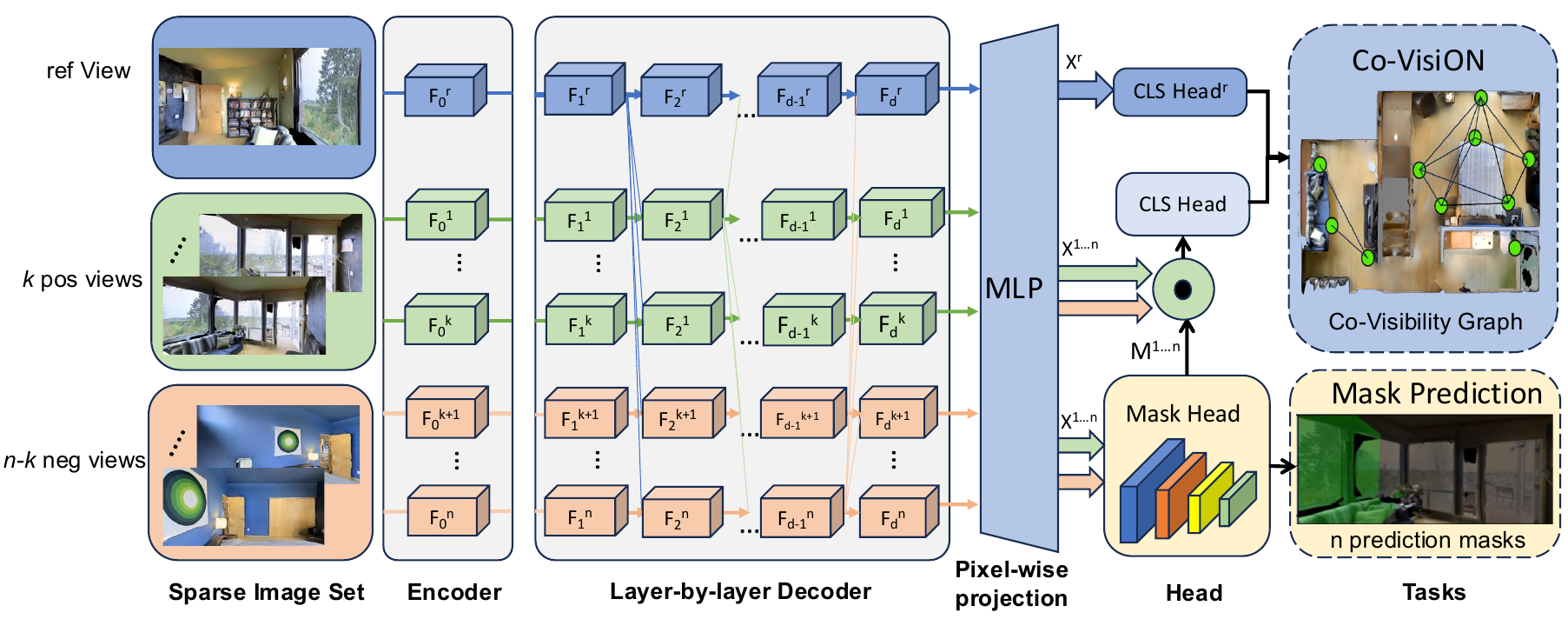}
    \vspace{-3mm}
    \caption{\textbf{Multi-view Covis pipeline.} Multi-view Covis is built on the MV-DUSt3R~\cite{tang2024mv} encoder-decoder backbone, following the same attention pattern across views. The encoder produces initial features \(F_0\) for each view, and the decoder refines these features layer-by-layer, producing intermediate outputs \(F_1, \ldots, F_{d-1}\). Decoder outputs $F_d$ are projected to pixel-wise features $X$ via MLP, then subsequently filtered by a mask $M$ to obtain $X'$. Visual tokens for the reference, positive, and negative views are shown in blue, green, and orange respectively. Gray and bold arrows indicate cross-view attention in the decoder. The reference (anchor) uses a separate prediction head, while the positive and negative views share the classification (CLS) and mask heads. }

    \label{fig:pipeline}
    \vspace{-6mm}
\end{figure*}
\vspace{-0.2em}
\subsection{Naive Baselines}
\vspace{-0.2em}
\label{sec:baseline}

We first benchmark several conventional approaches for Co-VisiON, where inference involves exhaustive pairwise comparisons between all node pairs to generate a co-visibility map. This includes binary classification, supervised contrastive learning and feature matching.

 \noindent\textbf{Pairwise supervised classification.} We evaluate three classic backbones—ViT~\cite{dosovitskiy2020vit}, VGG~\cite{simonyan2014vgg}, and ResNet~\cite{he2016resnet}—for pairwise co-visibility prediction. Each model extracts image pair features and uses a classification head to predict binary probabilities, trained jointly. During inference, predictions are thresholded to form a binary adjacency matrix. AUC is computed across thresholds from 0 to 1 (detailed in~\cref{dataset_metric}).

\noindent\textbf{Contrastive learning approaches.} We test two contrastive approaches. The first uses a ResNet backbone with InfoNCE loss~\cite{oord2018infoNCE}, following SimCLR~\cite{chen2020simple}, but with supervised contrastive learning: co-visible image pairs are positives, others are negatives. The model is fine-tuned from pretrained SimCLR.
The second uses NetVLAD~\cite{arandjelovic2016netvlad}, which combines a ResNet backbone with a VLAD aggregation layer to extract global descriptors. For both methods, pairwise feature distance (normalized to [0, 1]) is used to predict edge probabilities in the co-visibility graph.

\noindent\textbf{Image feature matching.} SIFT+RANSAC~\cite{fischler1981ransac} and SuperGlue~\cite{sarlin2020superglue} serve as classic baselines for image feature matching. SuperGlue uses a graph neural network to match keypoints between image pairs, whereas SIFT+RANSAC relies on traditional feature matching with outlier rejection. The percentage of matches compared to the maximum possible matches is treated as the edge probability for the co-visibility graph.

%% file: sec/5_methods.tex
\vspace{-1mm}
\subsection{Improved Baselines}
\vspace{-1mm}

\noindent Beyond conventional baselines, we explore whether recent vision models can perform better on Co-VisiON, including state-of-the-art 3D stereo reconstruction methods that can inherently derive co-visibility, and large Vision Language Models (VLMs) that excel in visual understanding. More importantly, we propose and test a new baseline--Covis, leveraging multi-view masked classification.

\begin{figure}[b]
    \centering
    \vspace{-5mm}
    \includegraphics[width=0.48\textwidth]{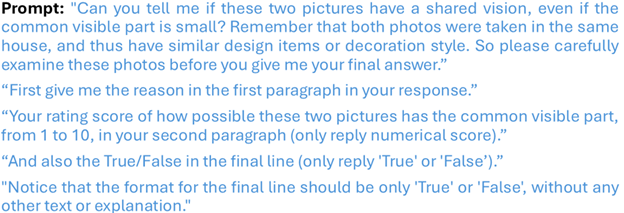}
    \vspace{-6mm}

    \caption{An example of the prompt used in our experiments. A unified prompt is adopted across all vision-language models.}
    \vspace{-3mm}
    \label{fig:prompt_vlm}
\end{figure}

\noindent \textbf{Large VLMs for Co-VisiON.} We evaluate the visual reasoning capabilities of general-purpose VLMs, including GPT-4o~\cite{openai2023gpt4}, Gemini 2.0 Flash~\cite{team2023gemini}, and Qwen2.5-VL 72b~\cite{yang2024qwen2}. Each model takes as input an image pair along with a prompt that explicitly asks it to reason about the co-visibility between the two views. In addition, we employ SpatialRGPT~\cite{NEURIPS2024_f38cb4cf}, a model specifically tailored for spatial reasoning tasks. SpatialRGPT builds on VILA~\cite{lin2024vila}, which natively supports multi-image input. We extend it to process image pairs with a shared visual encoder and a fine-tuned LLaMA3-70B~\cite{llama3modelcard} decoder for co-visibility reasoning. We use the same prompt for all VLM methods, as shown in~\cref{fig:prompt_vlm}. These models leverage their pre-trained reasoning capabilities to compare the images, compute pairwise co-visibility scores, and generate a co-visibility graph.

\noindent \textbf{3D Reconstruction models for Co-VisiON.} We use the DUSt3R~\cite{wang2023dust3r} and MV-DUSt3R+~\cite{tang2024mv} frameworks to generate global point clouds from input photos. The presence of binary topology edges is determined by the overlap of these point clouds. This approach is counterintuitive, as 3D reconstruction typically relies on topology map node relationships. Instead, we use 3D reconstruction to generate the co-visibility graph to highlight its limitations, including domain gap issues and frequent misjudgments of negative samples shown in~\cref{tab:table1}.

\vspace{-2mm}
\subsection{Covis}
By analyzing human evaluation patterns, we design \textbf{\textit{Covis}}, a multi-view encoder-decoder pipeline for co-visibility prediction, shown in~\cref{fig:pipeline}. To validate the advantages of multi-view reasoning, we perform an ablation study comparing Covis with its pairwise version. This helps assess the benefits of leveraging multiple views.

\noindent \textbf{Architecture.} CroCov2~\cite{Weinzaepfel_2023_ICCV} serves as the backbone for both pairwise and multi-view Covis, trained with a BCE loss to predict co-visibility. Experiments show that freezing the backbone during training yields the best results; this variant is referred to as \textbf{Covis-freeze} in subsequent discussions. In the pairwise setting, a reference image is paired with one other viewpoint to predict co-visibility. The multi-view Covis extends this by using the MV-DUSt3R~\cite{tang2024mv} encoder-decoder, enabling information exchange across multiple views for richer spatial reasoning. During training, positive and negative pairs are sampled from the ground truth co-visibility graph, while inference uses mixed pairs from the same scene to aid pairwise prediction. This multi-view design captures broader context by processing multiple images simultaneously.

\noindent \textbf{Mask prediction.} Non-co-visible information contaminates the reference image’s representation, hindering positive view learning (see Appendix). To filter out irrelevant features, we introduce a learnable mask $\mathbf{M}^i \in \mathbb{R}^{H \times W}$ for each view $i$, computed as $\mathbf{M}^i = \operatorname{conv}(\mathbf{X}^i)$,  
where $\mathbf{X}^i \in \mathbb{R}^{H \times W \times C}$ is a pixel-wise feature map projected from the corresponding decoder token $\mathbf{F}^i_d \in \mathbb{R}^T$ through a shared MLP: $\mathbf{X}^i = \operatorname{MLP}(\mathbf{F}^i_d)$. 
The MLP learns to expand 1D token into a 2D spatial feature map and is shared across all views except the reference view.

We then apply the mask to $\mathbf{X}$ via pixel-wise multiplication to preserve relevant regions from positive views:
$\mathbf{X}^{\prime}=\mathbf{X} \odot\mathbf{M}$, where $\mathbf{X}^{\prime}$ denotes the masked feature map, and $\odot$ denotes pixel-wise multiplication.
Both the co-visibility prediction and mask prediction tasks use BCE loss but serve distinct purposes: the former predicts spatial relationships based on flattened 1D global features for image pairs, while the latter supervises pixel-wise co-visible regions, providing stronger localized signals.

\vspace{-1mm}
\begin{table}[ht]

\vspace{-1mm}
\caption{Comparison of different mask types with varying noise levels in Covis' test performance (IoU means Graph IoU).}
\centering
\vspace{-3mm}
\scriptsize
\resizebox{0.44\textwidth}{!}{
\begin{tabular}{l|c|cc|cc}
\toprule
\textbf{Mask Type} & \textbf{Noise} & \multicolumn{2}{c|}{\textbf{Gibson}} & \multicolumn{2}{c}{\textbf{HM3D}} \\
 &  & IoU & AUC & IoU & AUC \\
\hline
No Mask & - & 0.52 & 0.49 & 0.48 & 0.42 \\
Supervised Mask & None & 0.59 & 0.57 & 0.57 & 0.56 \\
Supervised Mask & Low & 0.59 & 0.56 & 0.59 & 0.55 \\
Supervised Mask & High & 0.56 & 0.50 & 0.50 & 0.48 \\
Applied GT Mask & - & 0.74 & 0.72 & 0.71 & 0.69 \\
\bottomrule
\end{tabular}
}
\vspace{-2mm}
\label{tab:mvbce_mask_comparison}
\end{table}

\begin{figure*}[h]
    \centering
    \begin{subfigure}[b]{0.45\textwidth}
        \centering
        \includegraphics[width=\textwidth]{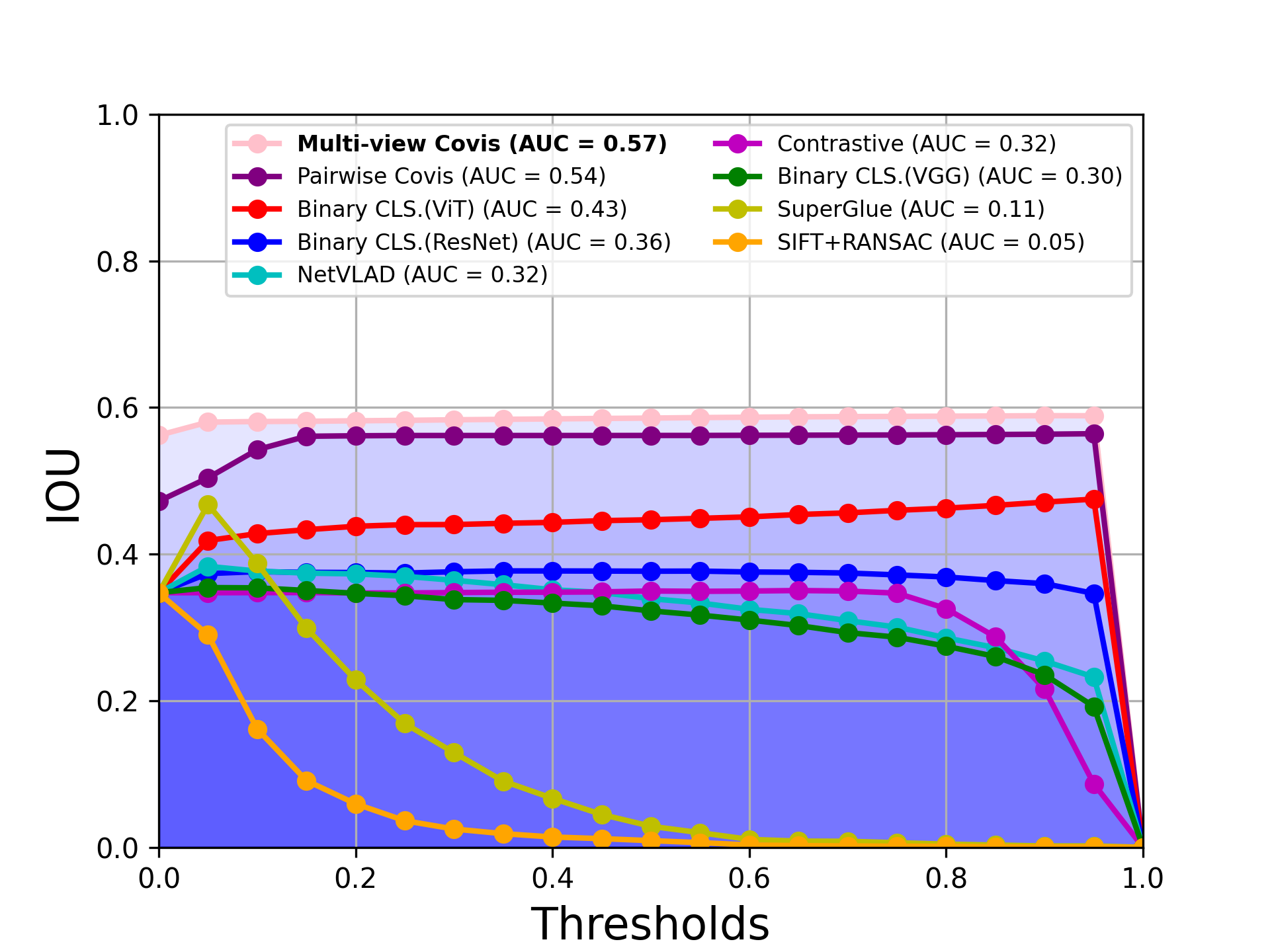}
        \caption{IoUs for Gibson dataset}
    \end{subfigure}
    \begin{subfigure}[b]{0.45\textwidth}
        \centering
        \includegraphics[width=\textwidth]{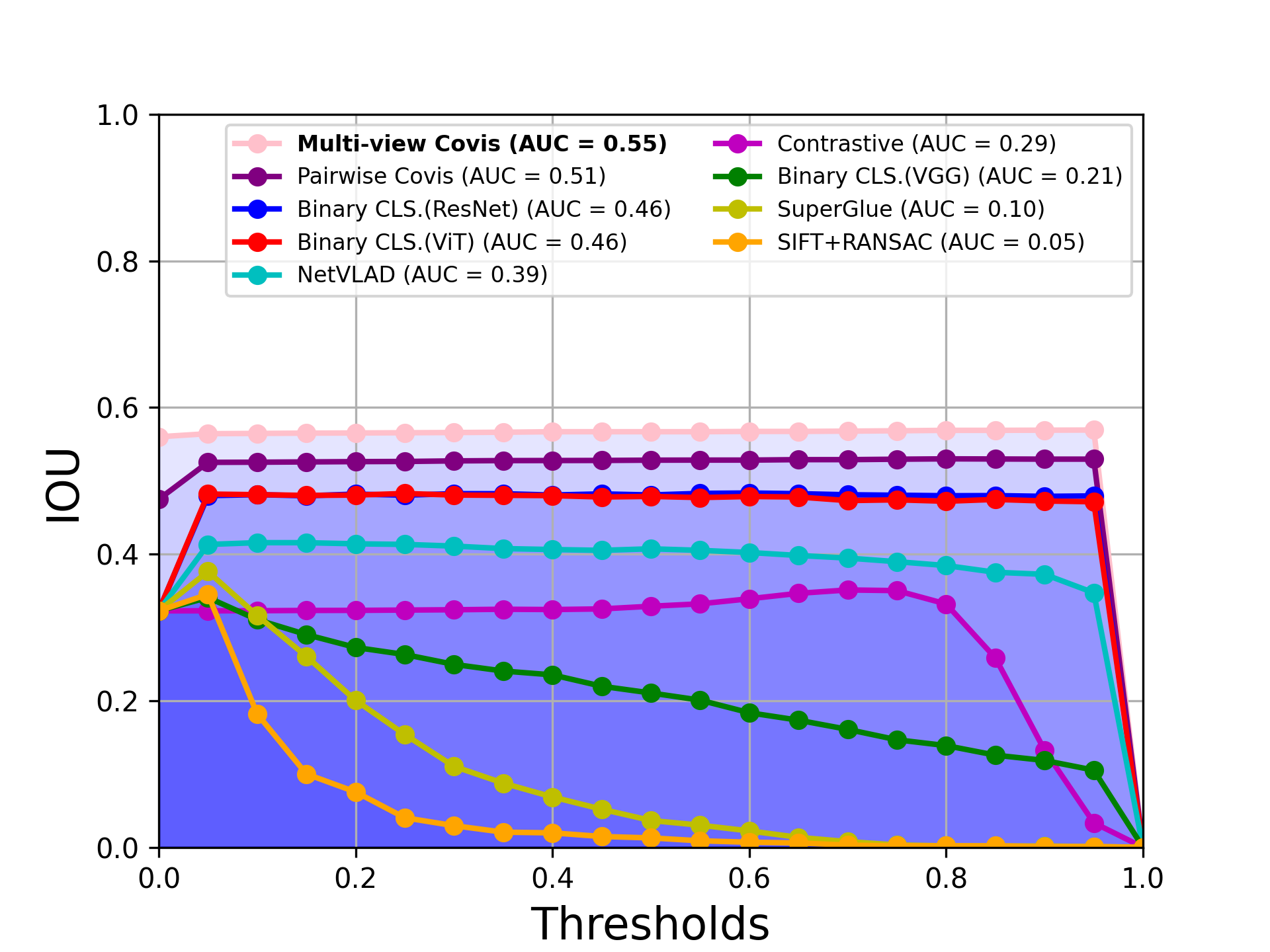}
        \caption{IoUs for HM3D dataset}
    \end{subfigure}
    \hspace{10pt}
    \vspace{-4mm}
    \caption{Comparison of IoU scores across various thresholds for the top-performing vision models on the Gibson and HM3D datasets.}
    \label{fig:AUC}
\end{figure*}
\vspace{-3.0mm}
\section{Discussion} 
Overall, our experiments demonstrate that GPT-4o and CoVis consistently outperform both naive and enhanced baselines across diverse datasets. In this section, we first discuss key findings observed throughout the experiments, followed by an analysis of common failure modes to identify potential avenues for improvement.

\subsection{Key findings}
\noindent\textbf{Summary of Naive baseline performance.} We present results in~\cref{tab:table1} and AUC curves in~\cref{fig:AUC}. Among conventional baselines, Vision Transformer (ViT) achieves the most stable IoUs across thresholds, with contrastive learning showing a similar trend.
Supervised methods, especially ViT, outperform feature-matching approaches on both datasets. Contrastive learning, though trailing supervised methods, still outperforms feature-matching by effectively managing positive and negative pairs—highlighting its potential for co-visibility reasoning.
All conventional baselines fall significantly short of human performance. Pairwise matching and classification methods alone are insufficient for Co-VisiON.

\vspace{-1.5mm}
\begin{table}[h]
\centering
\small
\setlength{\tabcolsep}{4pt}
\renewcommand{\arraystretch}{0.9}
\vspace{-2mm}
\caption{
Best IoU performance across different difficulty levels. The top-3 performers under each difficulty level are color-coded.
}
\vspace{-2mm}
\resizebox{0.46\textwidth}{!}{ 
\begin{tabular}{llcccc}
\toprule
\textbf{Setting} & \textbf{Method} & \textbf{Easy} & \textbf{Med.} & \textbf{Hard} & \textbf{Avg.} \\
\midrule
\multirow{6}{*}{\shortstack{Image Overlap \\ (edge level)}} 
& GPT-4o         & \textcolor{blue}{0.97} & \textcolor{blue}{0.92} & \textcolor{red}{\textbf{0.34}} & \textcolor{red}{\textbf{0.63}} \\
& Gemini-2.0-Flash       & \textcolor{red}{\textbf{1.00}} & \textcolor{teal}{0.99} & 0.14 & 0.42 \\
& Qwen2.5-VL 72b           & \textcolor{red}{\textbf{1.00}} & 
\textcolor{teal}{0.99} & 0.14 & 0.41 \\
& SpatialRGPT         & \textcolor{red}{\textbf{1.00}} & \textcolor{teal}{0.99} & 0.13 & 0.35 \\
& Covis          & \textcolor{teal}{0.99} & 0.88 & \textcolor{blue}{0.24} & \textcolor{blue}{0.59} \\
& Covis (freeze) & \textcolor{red}{\textbf{1.00}} & 0.89 & \textcolor{teal}{0.30} & \textcolor{teal}{0.61} \\
\midrule
\multirow{6}{*}{\shortstack{Scene Sparsity \\ (graph level)}} 
& GPT-4o         & \textcolor{red}{\textbf{0.83}} & \textcolor{teal}{0.64} & \textcolor{red}{\textbf{0.57}} & \textcolor{red}{\textbf{0.63}} \\
& Gemini-2.0-Flash       & 0.75 & 0.39 & 0.28 & 0.42 \\
& Qwen2.5-VL 72b          & 0.77 & 0.40 & 0.28 & 0.41 \\
& SpatialRGPT         & 0.72 & 0.38 & 0.26 & 0.35 \\
& Covis          & \textcolor{blue}{0.80} & \textcolor{blue}{0.63} & \textcolor{blue}{0.52} & \textcolor{blue}{0.59} \\
& Covis (freeze) & \textcolor{teal}{0.81} & \textcolor{red}{\textbf{0.65}} & \textcolor{teal}{0.54} & \textcolor{teal}{0.61} \\
\bottomrule
\end{tabular}
}
\vspace{-3mm}
\label{tab:categorized}
\end{table}

\begin{table*}[t]
\caption{\textbf{IoU* (best IoU) and AUC results on the Gibson and HM3D datasets.} We compare baseline and manual performance on the test sets. \textbf{Bold} values indicate the best performance in each category. The top three performances are color-coded: \textcolor{red}{red} for the best, \textcolor{teal}{teal} for second, and \textcolor{blue}{blue} for third. For 3D reconstruction, VLM categories and human annotation, AUC values are equivalent to IoU since they are not affected by thresholding. Binary CLS. stands for binary classification. Human annotation performance represents the upper bound.}
\centering
\resizebox{0.85\textwidth}{!}{
\begin{tabular}{l c c c c c c c} 
\toprule[1.0pt]

\multirow{2}{*}{\textbf{Category}} & \multirow{2}{*}{\textbf{Method}} & \multirow{2}{*}{\textbf{Backbone}} & \multirow{2}{*}{\makecell{\textbf{Pairwise(\ding{55}) /} \\ \textbf{Multiview(\ding{51})}}} & \multicolumn{2}{c}{\textbf{Gibson}} & \multicolumn{2}{c}{\textbf{HM3D}} \\ 
\cline{5-8} 
 &  &  &  & \textbf{IoU* ($\uparrow$)} & \textbf{AUC ($\uparrow$)} & \textbf{IoU* ($\uparrow$)} & \textbf{AUC ($\uparrow$)} \\ 
\hline

\multirow{2}{*}{\textbf{Feature Matching}} 
& \multicolumn{2}{c}{\textit{SuperGlue}~\cite{sarlin2020superglue} }   & \ding{55} & \textbf{0.47} & \textbf{0.11} & \textbf{0.38} & \textbf{0.10} \\ 
& \multicolumn{2}{c}{\textit{SIFT}~\cite{lowe2004sift} + \textit{RANSAC}~\cite{fischler1981ransac}  } & \ding{55} & 0.35 & 0.05 & 0.34 & 0.05 \\ 
\cmidrule{1-8}

\multirow{2}{*}{\textbf{Contrastive}}  
& \textit{-}~ & ResNet18~\cite{he2016resnet} & \ding{55} & 0.35 & \textbf{0.33} & 0.35 & 0.29 \\ 
& \textit{NetVLAD}~\cite{arandjelovic2016netvlad} & ResNet18~\cite{he2016resnet} & \ding{55} & \textbf{0.38} & 0.32 & \textbf{0.42} & \textbf{0.39} \\ 
\cmidrule{1-8}

\multirow{2}{*}{\textbf{3D Reconstruction}}  
& \textit{DUSt3R}~\cite{wang2023dust3r} & CroCo v2~\cite{Weinzaepfel_2023_ICCV} & \ding{55} & 0.54 & 0.54 & 0.40 & 0.40 \\ 
& \textit{MV-DUSt3R+}~\cite{tang2024mv} & CroCo v2~\cite{Weinzaepfel_2023_ICCV} & \ding{55} & \textbf{0.56} & \textbf{0.56} & \textbf{0.45} & \textbf{0.45} \\ 
\cmidrule{1-8}

\multirow{5}{*}{\textbf{Binary CLS.}} 
& -  & ViT~\cite{dosovitskiy2020vit}& \ding{55} & 0.47 & 0.43 & \textbf{0.48} & \textbf{0.46} \\ 
& - & VGG~\cite{simonyan2014vgg} & \ding{55} & 0.35 & 0.30 & 0.34 & 0.21  \\ 
& - & ResNet18~\cite{he2016resnet} & \ding{55} & 0.38 & 0.36 & \textbf{0.48} & \textbf{0.46} \\ 
& - & CroCo v2~\cite{Weinzaepfel_2023_ICCV} & \ding{55} & \textbf{0.50} & \textbf{0.45} & 0.43 & 0.38 \\ 
& - & CroCo v2~\cite{Weinzaepfel_2023_ICCV} & \ding{51} & 0.48 & 0.41 & 0.41 & 0.37 \\ 
\cmidrule{1-8}

\multirow{2}{*}{\textbf{Masked Binary CLS.}}
& \textit{Covis (Ours)}  & CroCo v2~\cite{Weinzaepfel_2023_ICCV} & \ding{55} & 0.56 & 0.54 & 0.53 & 0.51 \\ 
& \textit{Covis (Ours)} & CroCo v2~\cite{Weinzaepfel_2023_ICCV} & \ding{51} & \textbf{\textcolor{blue}{0.59}} & \textbf{\textcolor{blue}{0.57}} & \textbf{\textcolor{blue}{0.57}} & \textbf{\textcolor{teal}{0.56}} \\ 
& \textit{Covis-freeze (Ours)} & CroCo v2~\cite{Weinzaepfel_2023_ICCV} & \ding{51} & \textbf{\textcolor{teal}{0.61}} & \textbf{\textcolor{blue}{0.57}} & \textbf{\textcolor{teal}{0.58}} & \textbf{\textcolor{teal}{0.56}} \\ 
\cmidrule{1-8}

\multirow{4}{*}{\textbf{VLM}}
& \multicolumn{2}{c}{\textit{Qwen2.5-VL 72b}~\cite{yang2024qwen2}} & \ding{55} & 0.41 & 0.41 & 0.39 & 0.39 \\
& \multicolumn{2}{c}{\textit{Gemini-2.0-Flash}~\cite{team2023gemini}} & \ding{55} & 0.42 & 0.42 & 0.39 & 0.39 \\
& \multicolumn{2}{c}{\textit{SpatialRGPT}~\cite{NEURIPS2024_f38cb4cf}} & \ding{55} & 0.49 & 0.49 & 0.37 & 0.37 \\

& \multicolumn{2}{c}{\textit{GPT-4o}~\cite{openai2023gpt4}} & \ding{51} & 0.58 & \textcolor{teal}{0.58} & 0.54 & \textcolor{blue}{0.54} \\

& \multicolumn{2}{c}{\textit{GPT-4o}~\cite{openai2023gpt4}} & \ding{55} & \textbf{\textcolor{red}{0.63}} & \textbf{\textcolor{red}{0.63}} & \textbf{\textcolor{red}{0.59}} & \textbf{\textcolor{red}{0.59}} \\

\cmidrule{1-8}

\multicolumn{3}{c}{\textbf{Human Annotation}} 
& \ding{51} & 0.72 & 0.72 & - & - \\ 

\bottomrule[1.0pt] 

\end{tabular}
 }
\vspace{-2mm}

\vspace{-4mm}
\label{tab:table1}
\end{table*}

\noindent\textbf{Pairwise vs multi-view performance.} Across both 3D reconstruction and Covis, we observe that multi-view approaches consistently outperform pairwise approaches by about 3\% in IoU, as shown in ~\cref{tab:table1}. Multi-view inputs provide richer spatial context, leading to more accurate 3D reconstruction and better feature learning. In contrast, pairwise Covis is limited by the information in a single image pair, resulting in reduced performance.

\noindent \textbf{The importance of masking.}
Multi-view methods generally outperform pairwise ones. However, without masking, Covis suffers a sharp performance drop as the negative sample ratio increases, particularly in multi-view scenarios. This is because more negative information is introduced, acting as distractors during training. More specifically, irrelevant positive and negative features exchanged during the encoder-decoder process lead to feature contamination. In~\cref{tab:mvbce_mask_comparison}, the learnable mask helps filter out irrelevant regions and consistently improves performance, achieving up to a ~10\% gain in challenging multi-view settings.

To further examine the effect of mask quality, we simulate noisy supervision by injecting boundary-localized Gaussian noise into ground truth masks. Boundaries are detected via Laplacian filtering; noise with standard deviation $\sigma \in {0.01, 0.5}$ is added and binarized at threshold $\tau = 0.5$, while preserving all original positives. Low-level noise slightly improves performance due to regularization, while high noise degrades it significantly, highlighting the importance of accurate masks.

\noindent \textbf{VLM performance.} We evaluate proprietary and open-source models without fine-tuning. \textit{GPT-4o} attains 63\% IoU-closest to human performance at 72\% and best in~\cref{tab:table1} - demonstrating superior 
co-visibility reasoning capability under sparse conditions. Conversely, open-source models \textit{Qwen2.5-VL 72b} and \textit{Gemini-2.0-Flash} perform worse, achieving just over 40\%, below many visual baselines. Moreover, as a spatial reasoning specialist, \textit{SpatialRGPT} scores under 50\%, highlighting VLM instability on novel tasks despite their visual understanding capabilities.

\vspace{-2mm}
\subsection{Failure Mode Analysis} 
\vspace{-1mm}
To further assess the robustness of each method, we conduct a failure mode analysis by categorizing the evaluation data along two dimensions: \textbf{image overlap} and \textbf{scene sparsity}. These correspond to different levels of sparsity in the underlying co-visibility graph: image overlap reflects \textit{edge-level sparsity} between image pairs, while scene sparsity captures \textit{graph-level sparsity} across the entire scene. This enables a more fine-grained understanding of model behavior when either the overlap between image pairs is low  or the overall connectivity of the scene graph is sparse.

In this experiment, we define three difficulty levels for each evaluation protocol:
\vspace{-1mm}
\begin{itemize}
  \item \textbf{Image Overlap}: overlap ratio per image pair — \textit{easy} $\geq 50\%$, \textit{medium} $10\% \leq \cdot < 50\%$, \textit{hard} $< 10\%$.
  \item \textbf{Scene Sparsity}: average pairwise overlap per scene — \textit{easy} $\geq 10\%$, \textit{medium} $4\% \leq \cdot < 10\%$, \textit{hard} $< 4\%$.
\end{itemize}
\vspace{-1mm}

\noindent\textit{Discussion.} In \cref{tab:categorized}, GPT-4o outperforms other VLMs. Both Covis and GPT-4o achieve the most significant improvements over all baselines in hard case scenarios. These hard cases highlight that limited viewpoint overlap and scene sparsity are the primary sources of difficulty.

%% file: sec/6_application.tex
\section{Applications of Co-VisiON}
\subsection{DUSt3R for scene reconstruction} \label{sec:dust3r}

\noindent \textbf{Co-VisiON guided 3D Reconstruction.} 
Topology maps guide 3D reconstruction by selecting pertinent frames within a specific area. For instance, when reconstructing a kitchen, only frames depicting the kitchen are considered relevant. In contrast, DUSt3R~\cite{wang2023dust3r} exhaustively uses all available frames regardless of spatial relevance, resulting in inefficiencies in both memory and computation. To improve the accuracy and efficiency of 3D reconstruction, we propose a sparse co-visibility graph and compare it against commonly used topologies such as random, star, complete, and the ground truth. Definitions of these graph structures are provided in the supplementary material. Specifically, we treat each frame as a reference and jointly optimize it only with the frames directly connected in the topology graph, thereby constraining the matching and reconstruction.

\begin{table}[ht]
\caption{Comparison of different graph structures within the DUSt3R framework. We evaluate the Average Pose Error (APE), along with time complexity (in seconds) and memory cost (in GB) per scene. Results are averaged over Gibson test set. The Random graph fails in 3D reconstruction test, making APE not applicable.}

\centering
\vspace{-2mm}
\small
\setlength{\tabcolsep}{3pt}
\resizebox{0.46\textwidth}{!}{
\begin{tabular}{l|c c c|c c} \toprule[1pt]
\textbf{Method} & \textbf{Random} & \textbf{Star} & \textbf{Covision} & \textbf{Complete} & \textbf{GT} \\ \hline
APE (meters)($\downarrow$) & Failed & 1.791 & 1.701 & 1.601 & 1.548 \\
Time Complexity($\downarrow$) & - & 23 & 21 & 32 & 25 \\
Memory Cost($\downarrow$) & - & 7.3 & 6.7 & 10.2 & 7.9 \\
\bottomrule[1pt] 
\end{tabular}
}
\vspace{-5mm}
\label{tab:comparison_table}
\end{table}

\noindent \textbf{Quantitative Analysis of Different Graphs.} To evaluate pose estimation and reconstruction quality, we calculate the average pose errors (APE) by matching estimated poses with their closest ground truth poses, as shown in~\cref{tab:comparison_table}. As expected, using complete and ground truth (GT) graphs yields lower APE, demonstrating more accurate 3D reconstruction. However, obtaining a ground truth graph and using the complete graph incurs high annotation and computational costs. Moreover, random and star graphs fail to produce a reliable 3D map, as shown in~\cref{fig:image_pairing_methods}. Therefore, we find the co-visibility graph to be one of the most suitable options for DUSt3R, balancing memory costs with a reliable 3D map, as shown in~\cref{fig:covision_dust3r}. Detailed reconstruction visualizations are available in the supplementary.

\begin{figure}[h]
    \centering
    
    \begin{subfigure}[b]{0.15\textwidth}
        \centering
        \includegraphics[width=\textwidth]{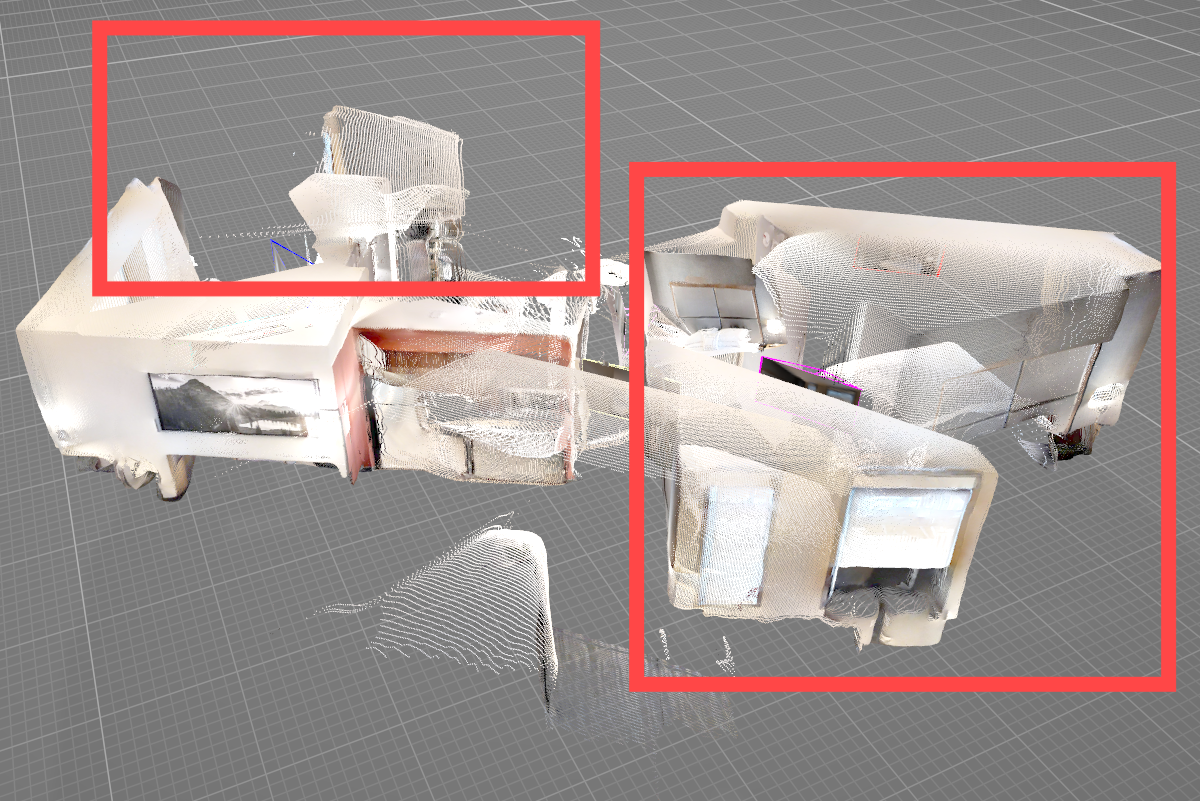}
        \caption{Star - Scene 1}
    \end{subfigure}
    \begin{subfigure}[b]{0.15\textwidth}
        \centering
        \includegraphics[width=\textwidth]{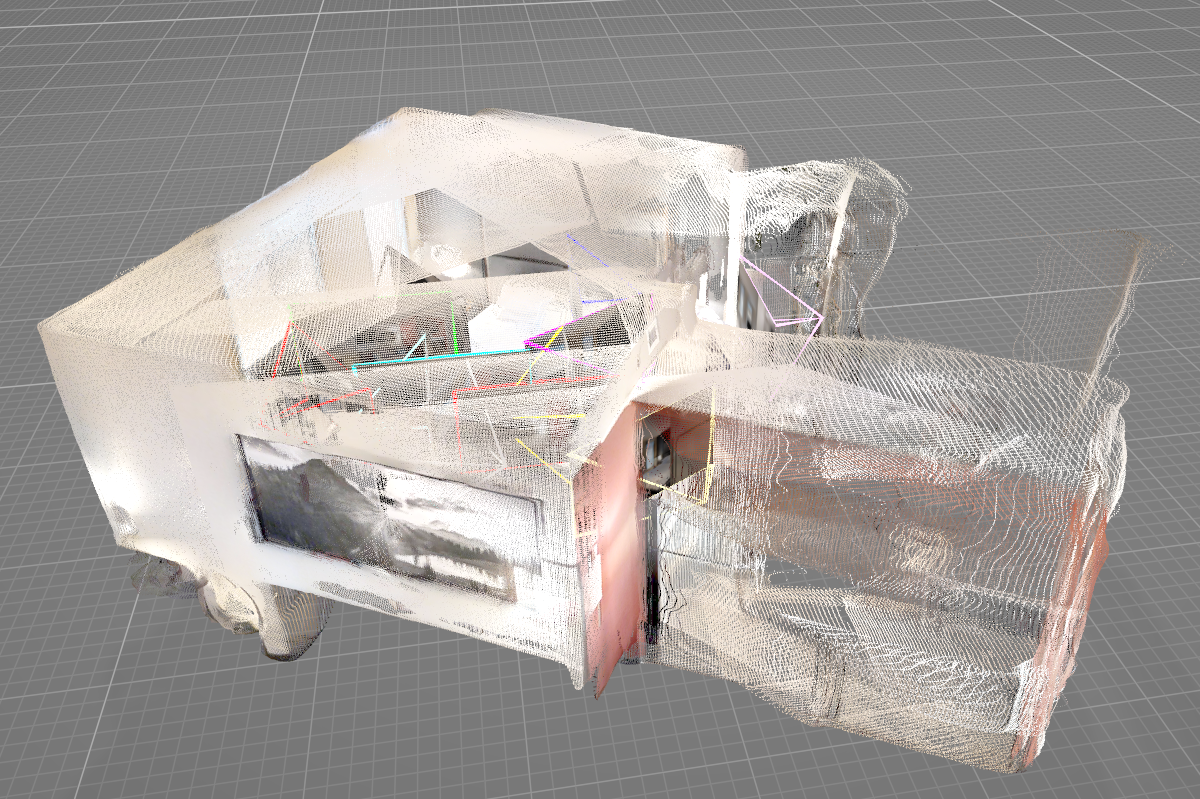}
        \caption{Covision - Scene 1}
    \end{subfigure} 
    \begin{subfigure}[b]{0.15\textwidth}
        \centering
        \includegraphics[width=\textwidth]{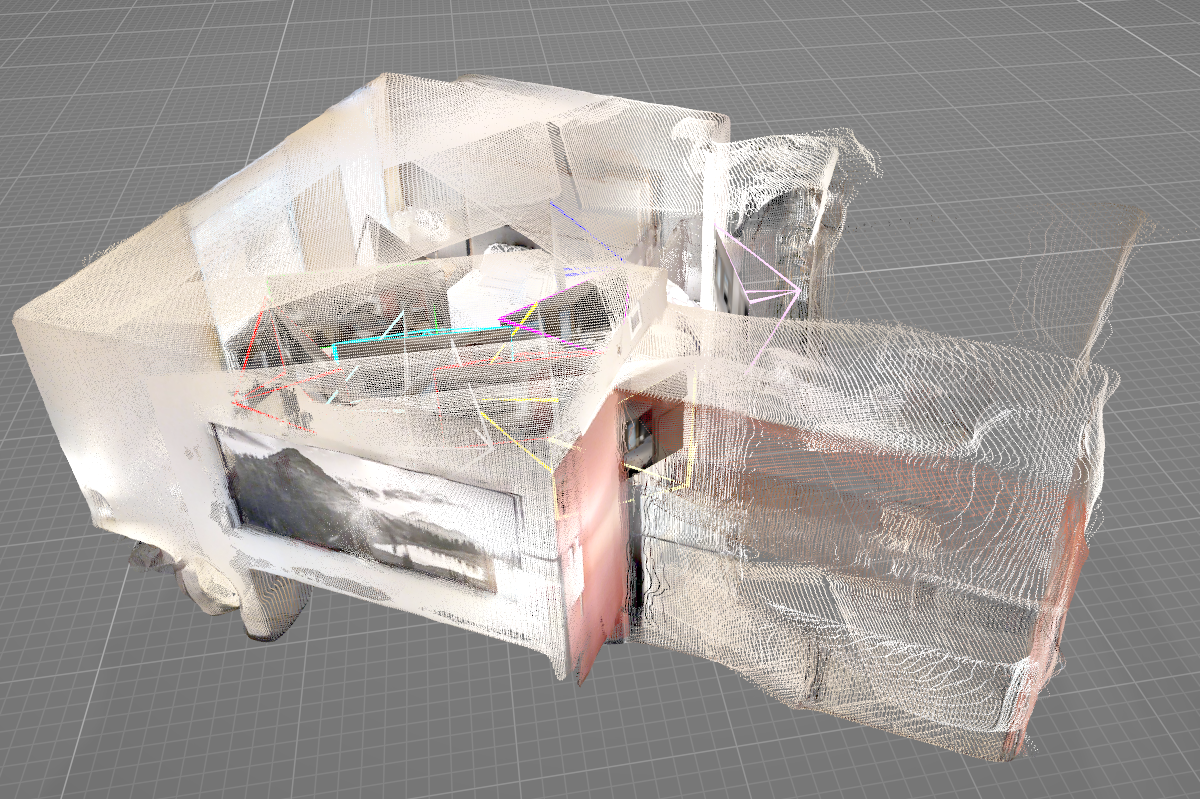}
        \caption{GT - Scene 1}
    \end{subfigure}

    \begin{subfigure}[b]{0.15\textwidth}
        \centering
        \includegraphics[width=\textwidth]{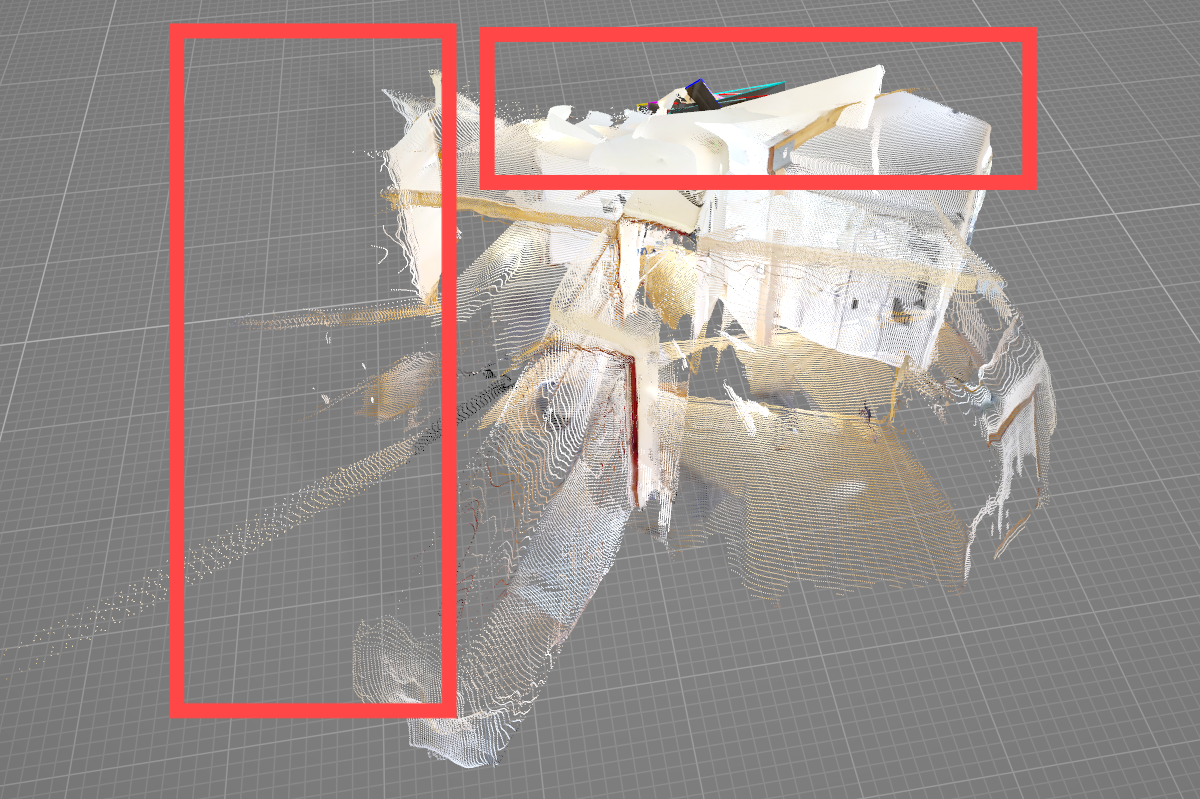}
        \caption{Star - Scene 2}
    \end{subfigure}
    \begin{subfigure}[b]{0.15\textwidth}
        \centering
        \includegraphics[width=\textwidth]{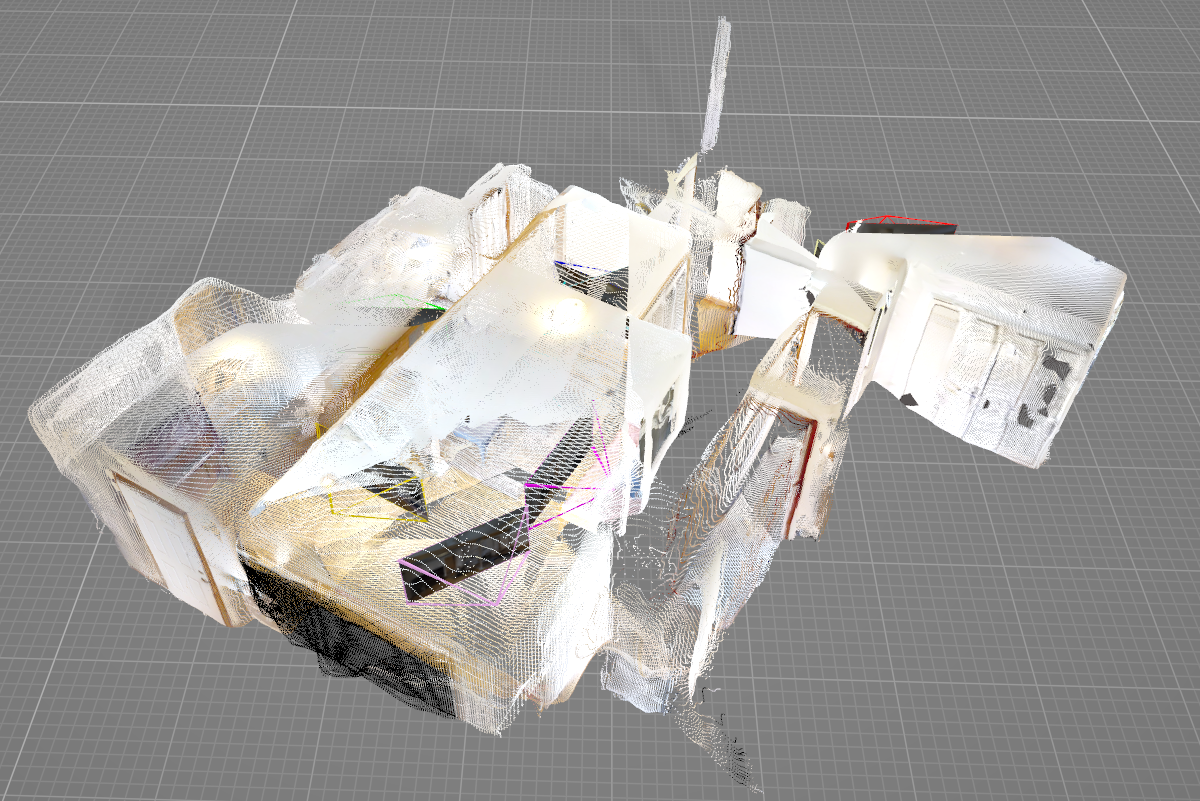}
        \caption{Covision - Scene 2}
    \end{subfigure}
    \begin{subfigure}[b]{0.15\textwidth}
        \centering
        \includegraphics[width=\textwidth]{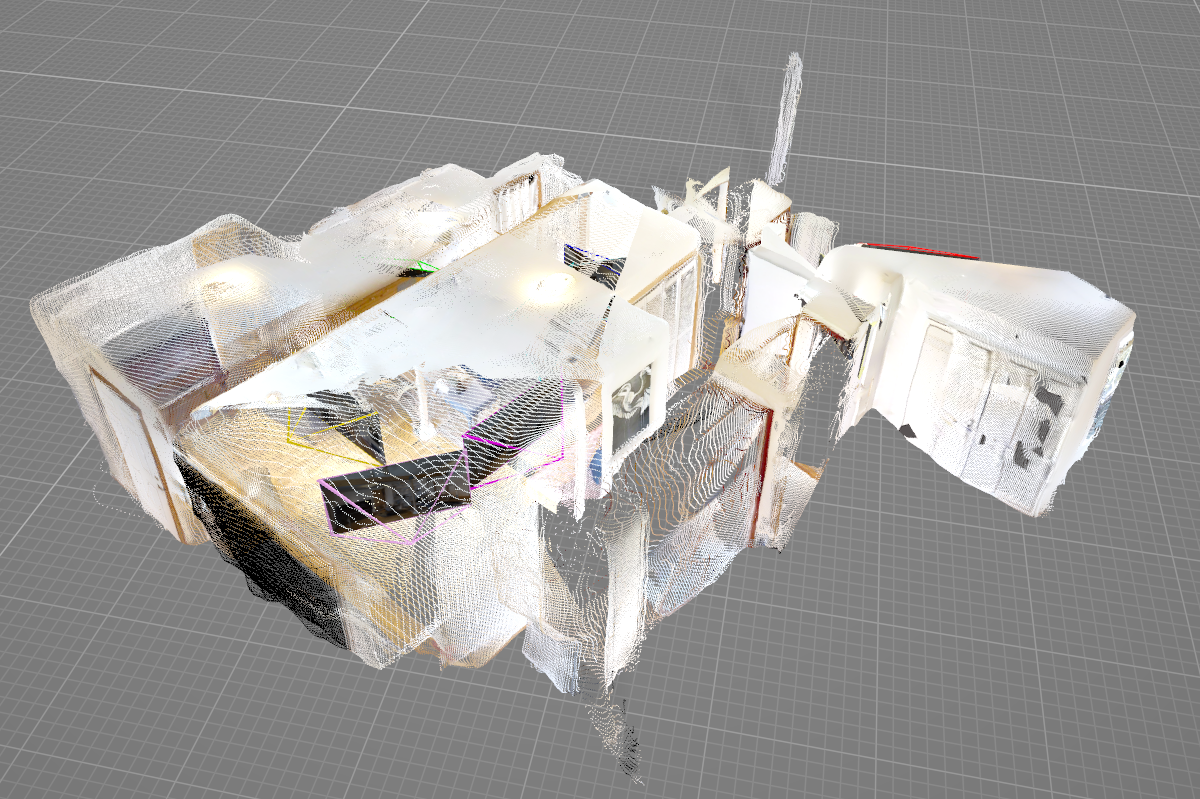}
        \caption{GT - Scene 2}
    \end{subfigure}
    \vspace{-2mm}
    \caption{3D reconstruction results by DUSt3R using co-visibility and GT graphs, showing similar performance.Regions highlighted by \textcolor{red}{red} boxes indicate reconstruction failures.}
    \vspace{-3mm}
    \label{fig:covision_dust3r}
\end{figure}

\vspace{-2mm}
\subsection{Automated Training Set Labeling} 

\noindent The co-visibility graph provides a topology for automatically labeling training data, such as selecting triplets in VPR, forming batches in 3D reconstruction, and pairing images for view completion. We illustrate this with CroCo~\cite{weinzaepfel2022croco}, a model that completes masked regions of a target image using a reference from a different viewpoint. 

\noindent To eliminate the need for manual annotation, we automate the process by leveraging graphs, where images are represented as nodes, and the edges encode co-visibility relationships based on 3D structure or estimated visibility. This approach allows for the automatic generation of training pairs by treating each image as a query and its connected neighbors as reference candidates.

\noindent We evaluate three strategies for generating CroCo training pairs: (1) a high co-visibility graph with over 50\% 3D surface overlap; (2) a sparse co-visibility graph from our method; and (3) a random graph. For each, masked regions are restricted to the co-visible area between query and reference, except in the random graph where masks are applied arbitrarily. This allows us to assess whether our sparse co-visibility graph is a more annotation-efficient alternative while maintaining performance.

\begin{table}[ht]
\vspace{-2mm}
\caption{PSNR for cross-view completion with different graphs.}
\centering
\vspace{-2mm}
\begin{tabular}{l|c c c} \toprule[1pt]
\textbf{Method} & \textbf{Random} & \textbf{Co-Vis} & \textbf{High Co-Vis} \\ \hline
PSNR ($\uparrow$) & 12.46 & \textbf{16.34} & 16.32 \\
\bottomrule[1pt]
\end{tabular}
\vspace{-3mm}
\label{tab:psnr_values}
\end{table}

\begin{figure}[h]
    \centering
    \begin{minipage}[t]{0.15\textwidth}
        \centering
        \includegraphics[width=\textwidth, height=1.5cm]{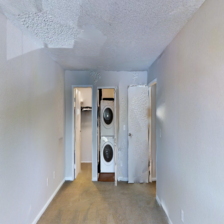}
        \vspace{-10pt}
    \end{minipage}
    \hspace{1pt}
    \begin{minipage}[t]{0.15\textwidth}
        \centering
        \includegraphics[width=\textwidth, height=1.5cm]{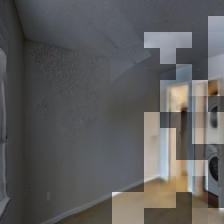}
        \vspace{-10pt}
    \end{minipage}
    \hspace{1pt}
    \begin{minipage}[t]{0.15\textwidth}
        \centering
        \includegraphics[width=\textwidth, height=1.5cm]{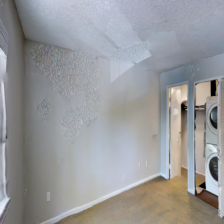}
        \vspace{-10pt}
    \end{minipage}

    \begin{minipage}[t]{0.15\textwidth}
        \centering
        \includegraphics[width=\textwidth, height=1.5cm]{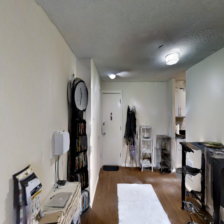}
        \vspace{-1pt}
        \small Reference
        \vspace{-3pt}
    \end{minipage}
    \hspace{1pt}
    \begin{minipage}[t]{0.15\textwidth}
        \centering
        \includegraphics[width=\textwidth, height=1.5cm]{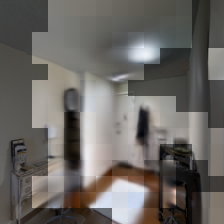}
        \vspace{-1pt}
        \small Prediction
        \vspace{-3pt}
    \end{minipage}
    \hspace{1pt}
    \begin{minipage}[t]{0.15\textwidth}
        \centering
        \includegraphics[width=\textwidth, height=1.5cm]{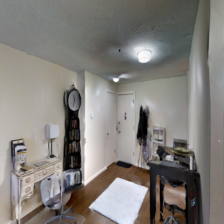}
        \vspace{-1pt}
        \small Target
        \vspace{-3pt}
    \end{minipage}
    \caption{Reconstructed co-visibility regions during inference. Visual comparison of different graph configurations. Top row: Co-visibility graph for training. Bottom row: High co-visibility graph for training. Columns: Reference image, predicted image, and ground truth target image. In the predicted images, the masked regions are visible as network input, and the brightened areas indicate the reconstructed output. Additionally, the random graph fails in cross-view completion.}

    \label{fig:covision_croco}
\end{figure}

\noindent\textbf{Quantitative Analysis on different graphs.} We conduct completion quality tests as shown in~\cref{fig:covision_croco}. Our results show that the sparse co-visibility graph achieves performance comparable to the high co-visibility graph as shown in~\cref{tab:psnr_values}, while being significantly easier to construct. This is because the sparse graph only requires selecting image pairs that meet minimal co-visibility criteria, whereas the high co-visibility graph demands stricter thresholds and more precise filtering, making it more computationally intensive and harder to scale. Moreover, it generalizes better to diverse scenes and supports more scalable training. These advantages make it a practical and efficient solution for automated training set labeling.
\vspace{-2mm}

%% file: sec/7_conclusion.tex
\vspace{-6mm}
\section{Conclusion}\label{sec:conclusion}

\noindent \textbf{{Limitation}.} Co-VisiON is limited to Gibson and HM3D scenes with limited human annotations. Future work will expand and diversify the dataset using additional sources, including real-world environments.

\noindent \textbf{{Summary}.}
This paper introduces the Co-Visibility reasONing (Co-VisiON) task, where models predict view connectivity and co-visibility graphs from sparse indoor image sets. We provide a curated dataset with human and automated annotations, show that Co-VisiON challenges existing baselines, and propose Covis - a multi-view model that surpasses pairwise methods by capturing complex spatial relationships. Co-VisiON aims to advance embodied scene understanding research.


%% file: sec/8_suppl.tex
\clearpage
\renewcommand{\thesection}{\Alph{section}}
\renewcommand{\thefigure}{\Roman{figure}}
\renewcommand{\thetable}{\Roman{table}}

\setcounter{section}{0}
\setcounter{figure}{0}
\setcounter{table}{0}
 \definecolor{greenpivot}{RGB}{0, 210, 0}

\section*{Appendix}
\begin{figure*}[ht!]
    \centering
    \includegraphics[width=0.96\textwidth]{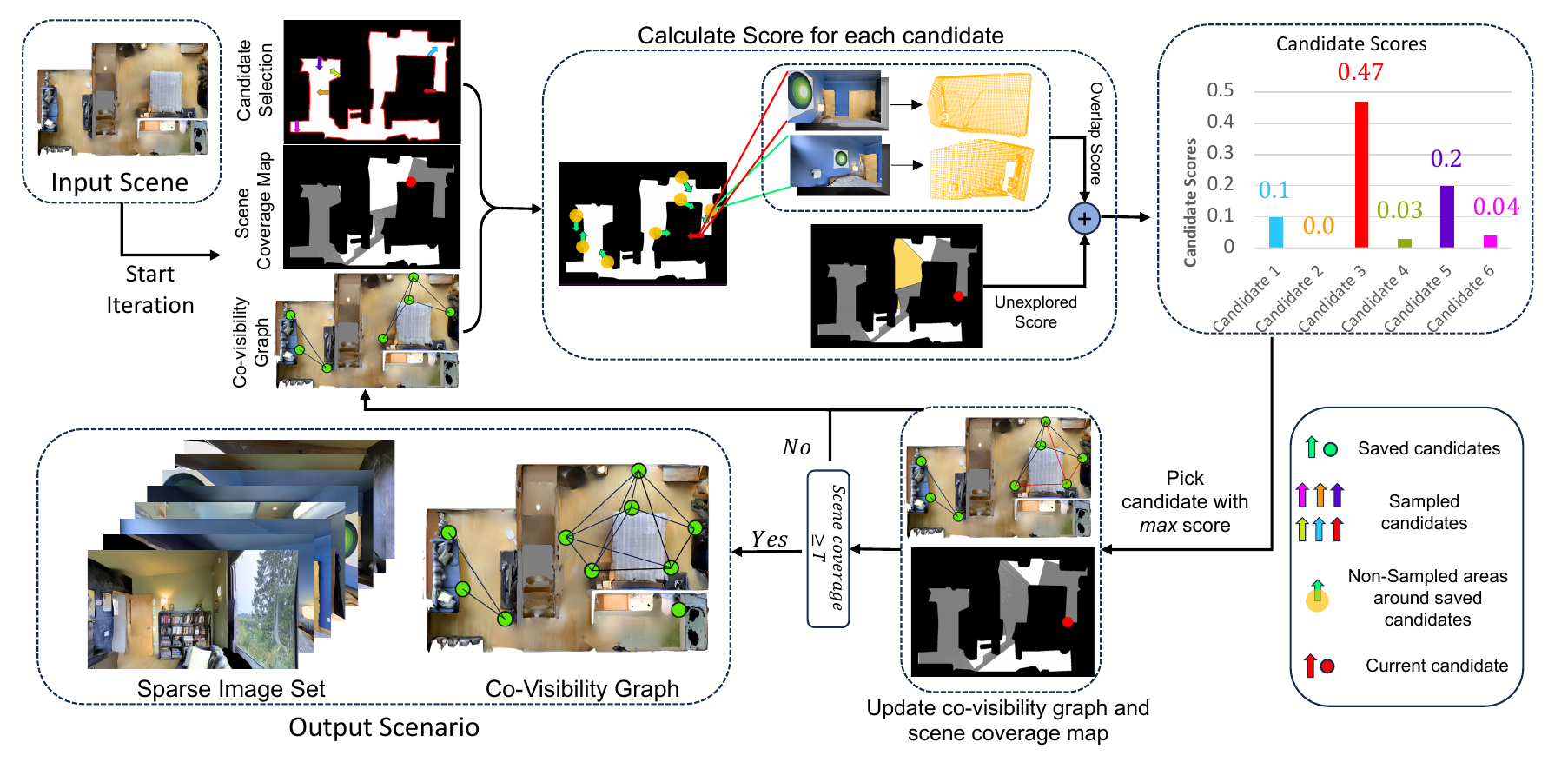}
    \caption{\textbf{Dataset generation:} Firstly, candidates are sampled from the possible locations shown as \textcolor{red}{red border} of the top-down map in candidate selection. Then the sampled candidates, along with a scene coverage map and the current co-visibility graph, are used to calculate scores for each candidate. The scores are computed with the \textcolor{greenpivot}{best saved candidates} from previous iterations. The top-scoring candidate among the currently sampled is chosen to update scene coverage and the co-visibility graph. This process is repeated until a scene coverage threshold is reached, at which point the best candidates' observations with a co-visibility graph are saved to create the dataset.} 
    \label{fig:workflow}
\end{figure*}

\noindent This supplementary provides additional details and results that could not fit in the main paper. Specifically, we include: (1) dataset and annotation generation (Gibson and HM3D), (2) top-down visualization of co-visibility graph, (3) evaluation metric details, (4) graph definitions used in DUSt3R and CroCo experiments, (5) ablation studies on Covis, and (6) Sim2Real downstream task details. All experiments and dataset generation were conducted on an HPC cluster with A100 and V100 GPUs, using a single GPU per run.

\begin{figure*}
    \centering
    \includegraphics[width=0.86\textwidth]{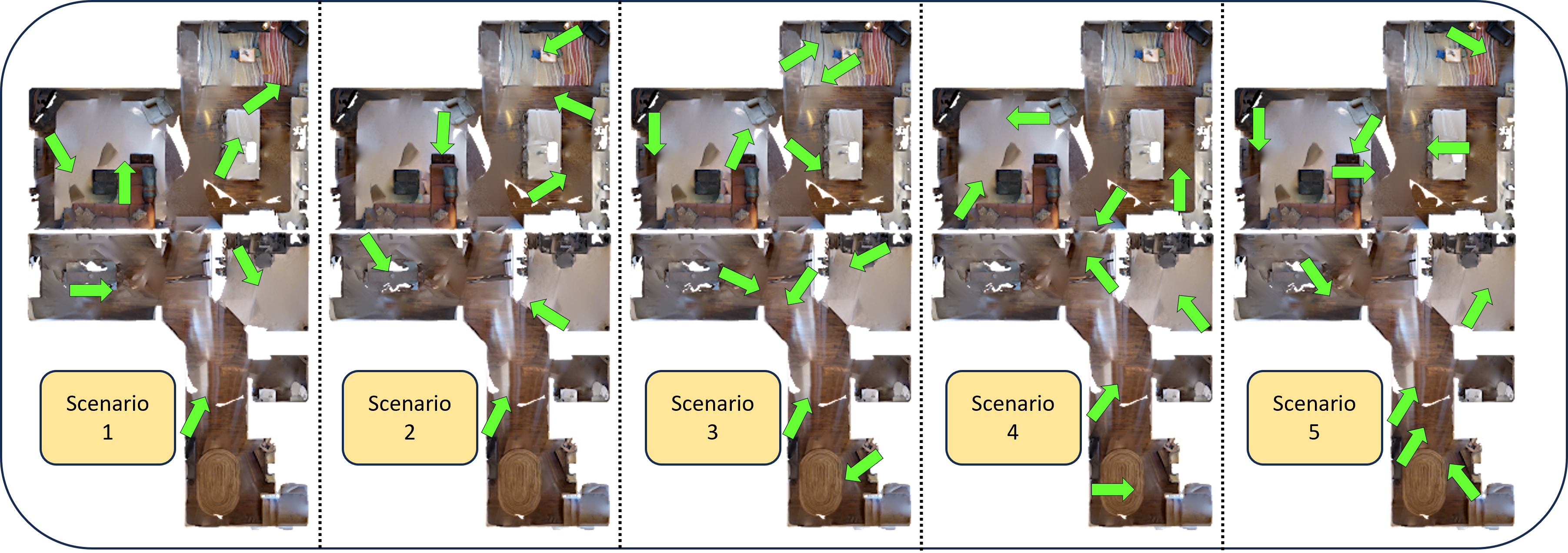}
    
    \caption{\textbf{We are able to generate different scenarios for a single scene. This is an example of 5 scenarios generated for Scene Goff}}\label{fig:scenario_scene}
\end{figure*}

\section{Dataset details}

\noindent We detail the dataset construction and annotation processes for the Gibson and HM3D environments, which form the basis of our co-visibility analysis. For each dataset, we define \textit{scenarios} as structured collections of images with associated camera poses, designed to ensure high coverage and annotation quality. Both datasets leverage simulated sensors (pinhole and stereo) to generate RGB and depth data, and we outline specific strategies used to handle scene diversity, pose recording, and structural alignment.

\noindent\textbf{Gibson}: This dataset covers 129 floors across 85 scenes. To ensure diversity, we randomize seed numbers during data generation, allowing for efficient image selection that maximizes spatial coverage with minimal redundancy. The image selection process is further detailed in~\cref{sec:annotation}. We record precise camera poses using Habitat-sim and organize images into structured scenarios for downstream use. 

\noindent\textbf{HM3D}: This dataset contains 755 annotated scenes. Since floor heights are not provided, we estimate them by clustering the Y-axis values of camera poses. This enables us to establish clear floor boundaries, which are essential for scene construction and co-visibility graph generation.

\section{Annotation generation details}
\subsection{Co-Visibility annotation overview}\label{sec:annotation}
As described above, with the 3D assets and depth information provided in the dataset, the ground truth annotation of co-visibility between two images should theoretically be straightforward. However, this process is more complex than it seems. We follow the dataset generation process outlined in~\cref{sec:dataset}. First, while placing cameras and selecting images, the image set must remain sparse, maintaining both co-visibility space and maximum scene coverage. Second, as a reasoning task, the annotations should align with human perception while being automatically applicable to large-scale datasets. To ensure annotation accuracy and consistency, we curate a smaller, human-annotated dataset, which serves as a human reasoning baseline for benchmarking Co-VisiON and validating the correctness of the automatically generated dataset.

\subsubsection{Automatic co-visibility annotation}\label{sec:automatic}
\textbf{Camera pose placing strategy.} While we could place cameras anywhere within the navigable areas of a scene, we select positions near walls or furniture to better mimic human photography behavior. This strategy aims to emulate real-world scenarios, where photographers often position themselves near peripheral areas to maximize coverage. As shown in ~\cref{fig:workflow}, the red-bordered regions between black obstacles and white navigable spaces indicate our preferred camera locations. The cameras are generally oriented away from walls to ensure comprehensive coverage, capturing a wide range of visual features that aid both neural network analysis and human interpretation.

\noindent\textbf{Image selection criterion.} First, to assess the scene coverage and the overlapping regions between any two images, we convert the depth image into a point cloud in global coordinates and use it in the subsequent scoring process. We select images in progressive iterations by adopting a scoring method to ensure we cover the entire scene with the fewest possible photos. In each iteration, $n_c$ randomly sampled candidates are generated based on the camera pose placing strategy where every candidate, $n_c^i$, possesses observations (RGB and Depth) along with the pose information of the agent and its sensors. 
We then choose the highest-scoring candidate based on the scoring function:
\begin{equation} \label{eq:2}
    S=\alpha \cdot O_u+\beta \cdot O_p,
\end{equation}
where $O_u$ represents the newly explored region covered by the projected point cloud, while $O_p$ denotes the previously explored region recorded from previous iterations. The $\alpha$ and $\beta$ are set to 0.9 and 0.1, respectively. This configuration reflects a preference for camera poses that explore more of the uncovered area.

\noindent After selecting the best candidate from $n_c$ possible candidates based on the weighted score as in \eqref{eq:2}, we then remove the positions near the best candidate (or within some radius $r$) from the viable candidate selection area using equation \eqref{eq:3}:
\begin{equation} \label{eq:3}
    d(p_i, p_s) <= r
\end{equation}
where $d(.)$ represents the Euclidean distance between the possible candidate location $p_i$ and the selected best candidate $p_s$. This particular pruning is shown as light-orange blobs in ~\cref{fig:workflow}. We observe that it is effective in covering the scene faster and encouraging sparse image sets. We iterate this data generation process until we explore more than 80$\%$ of the scene. All observations corresponding to the selected best candidates as well as their pairwise IOU are saved.

\subsubsection{Human co-visibility annotation}
\label{sec:human_annotation}
Besides developing the automatic annotation method, we also have human annotators manually label a subset of scenes. The human annotation mirrors how humans reason about spatial relationships and, therefore, serves two purposes: it helps assess the quality of automatic annotation, and more importantly, provides a human reasoning baseline for benchmarking Co-VisiON.

\noindent The process involves 6 scenes arbitrarily chosen from the automatically generated dataset from ~\cref{sec:automatic}. To facilitate human annotation, we develop a website with a GUI that loads pairs of images within the same scene for the trained annotators to determine if they can reason the co-visibility of any pair of images, and click to label them accordingly. Images are uniformly sampled from the same scenario to form a pair and presented to the annotators. 

Given images $I_a$, $I_b$ $\in \mathcal{I}$, the following criteria were considered for each pair:
\begin{itemize}
\item \textbf{Shared objects}:  If $I_a$ and $I_b$ share the view of the same objects such that an annotator can infer the two images' relative pose, they are labeled as connected.

\item \textbf{Object Continuity}: 

If \(I_a\) shows the left side of a sofa and \(I_b\) shows the right side, even if the sofa is not fully visible in either image, the partial views align in a way that the viewer can perceive them as continuous, and spatially reason the images as parts of the same object.
\item \textbf{Sub-Scene Relationship}: one image may be a more zoomed-in portion of the other. Or an image may have details obstructed by objects in its view.
\item \textbf{Featureless Surface}: If the overlapping region of $I_a$ and $I_b$ is devoid of distinguishable features---such as a plain wall---the pair is labelled as not connected. The absence of features hampers the ability to establish a clear co-visibility relationship between the images.
\end{itemize}

\noindent Along with the above criterion, this manual annotation process undergoes a stringent cross-validation phase where each scene is annotated at least twice by different annotators. When it is challenging to identify relationships in ambiguous scenarios, the images may be marked for further review. The resulting sets of annotations are then compared and discussed until a complete agreement is reached among the annotators. In this way, we ensure the annotations of spatial relationships are as precise and robust as possible.

We will release both the automatically labeled and the manually labeled datasets to facilitate future research.
The human annotation website will also be released so that anyone can extend the Co-VisiON task to a larger scale and for more diverse scenarios.

\section{Visualization of multiple scenarios from a single scene}
We can simulate different scenarios within the same scene by manipulating the seed, resulting in distinct image combinations. In this supplementary, we present five scenarios generated for Scene Goff through this approach shown in~\cref{fig:scenario_scene}.

\section{Evaluation metrics}
As discussed in~\cref{dataset_metric}, we use IoU (Intersection over Union) and AUC (Area Under Curve) as the evaluation criteria which are mathematically defined as following.
\begin{equation}
    \text{IOU}(\mathbf{A}, \hat{\mathbf{A}}) = \frac{\sum_{i}\sum_{j} \mathbf{A}_{ij} \land \hat{\mathbf{A}}_{ij}}{\sum_{i}\sum_{j} (\mathbf{A}_{ij} \lor \hat{\mathbf{A}}_{ij}) + \epsilon}
\end{equation}
where $\mathbf{A}_{ij}$ and $\hat{\mathbf{A}}_{ij}$ stand for the binary elements at the $i$th row and $j$th column of the corresponding matrices of the ground truth and predicted co-visibility graphs, $\mathcal{G}$ and $\hat{\mathcal{G}}$. The $\land$ and $\lor$ represent element-wise AND and OR operations respectively. A small constant $\epsilon$ is added to the denominator to avoid zero division.

Another metric that we use is Area Under Curve (AUC). Formally, suppose $\hat{\mathbf{A}}_{\tau}$ is the binary predicted matrix given the threshold $\tau$, and $\tau$ is sampled from a discrete and ordered set $\{t_1, t_2, \ldots, t_n\}$ from range $[0, 1]$. The AUC is computed as:
\begin{equation}
    \text{AUC} = \sum_{i=1}^{n-1} \frac{(\text{IOU}(\mathbf{A}, \hat{\mathbf{A}}_{t_{i+1}}) + \text{IOU}(\mathbf{A}, \hat{\mathbf{A}}_{t_i}))}{2} \cdot (t_{i+1} - t_i)
\end{equation}

\section{Details and graph definition in DUSt3R experiment}

To evaluate the performance of co-visibility prediction under different graph structures, we experiment with several types of graphs in the DUSt3R benchmark. Each graph is defined as follows:

\begin{itemize}
    \item \textbf{Complete graph} refers to a graph where every pair of images is directly connected. However, due to its resource-intensive nature, it becomes less practical for large-scale reconstructions.
    \item \textbf{Co-visibility graph} is resource-efficient and maintains geometric coherence, making it a viable alternative to the Complete strategy.
    \item \textbf{Star graph} consists of a central frame connecting all other images. This approach emphasizes a central perspective and proves valuable when a single viewpoint dominates the scene.
    \item \textbf{Ground Truth graph} connects a pair of images in its graph if they are within a predefined proximity. This graph relies on the absolute pose obtained from the ground truth for each pair of images.
\end{itemize}

\begin{figure*}
    \centering

    \begin{subfigure}[b]{0.22\textwidth}
        \centering
        \includegraphics[width=\textwidth]{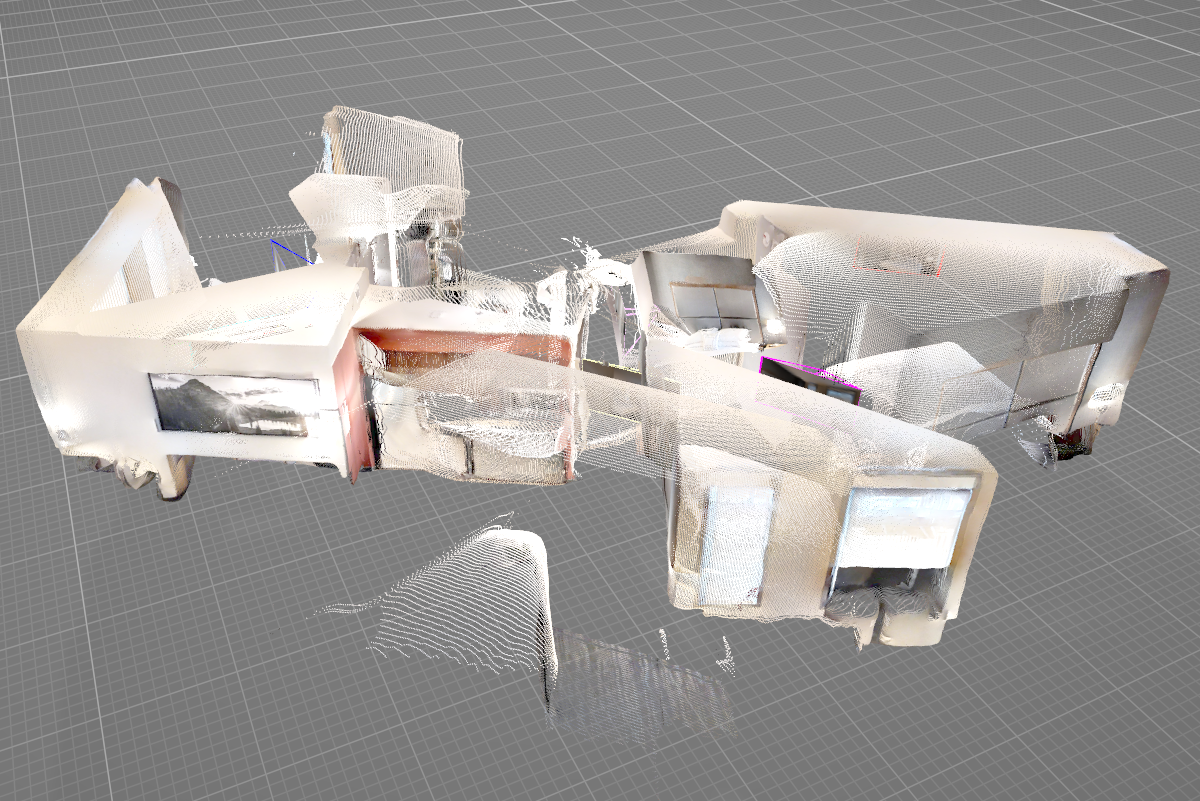}
        \caption{Star - Scene 1}
    \end{subfigure}
    \hspace{5pt}
    \begin{subfigure}[b]{0.22\textwidth}
        \centering
        \includegraphics[width=\textwidth]{figs/covision_scene6.png}
        \caption{Covision - Scene 1}
    \end{subfigure}
    \hspace{5pt}
    \begin{subfigure}[b]{0.22\textwidth}
        \centering
        \includegraphics[width=\textwidth]{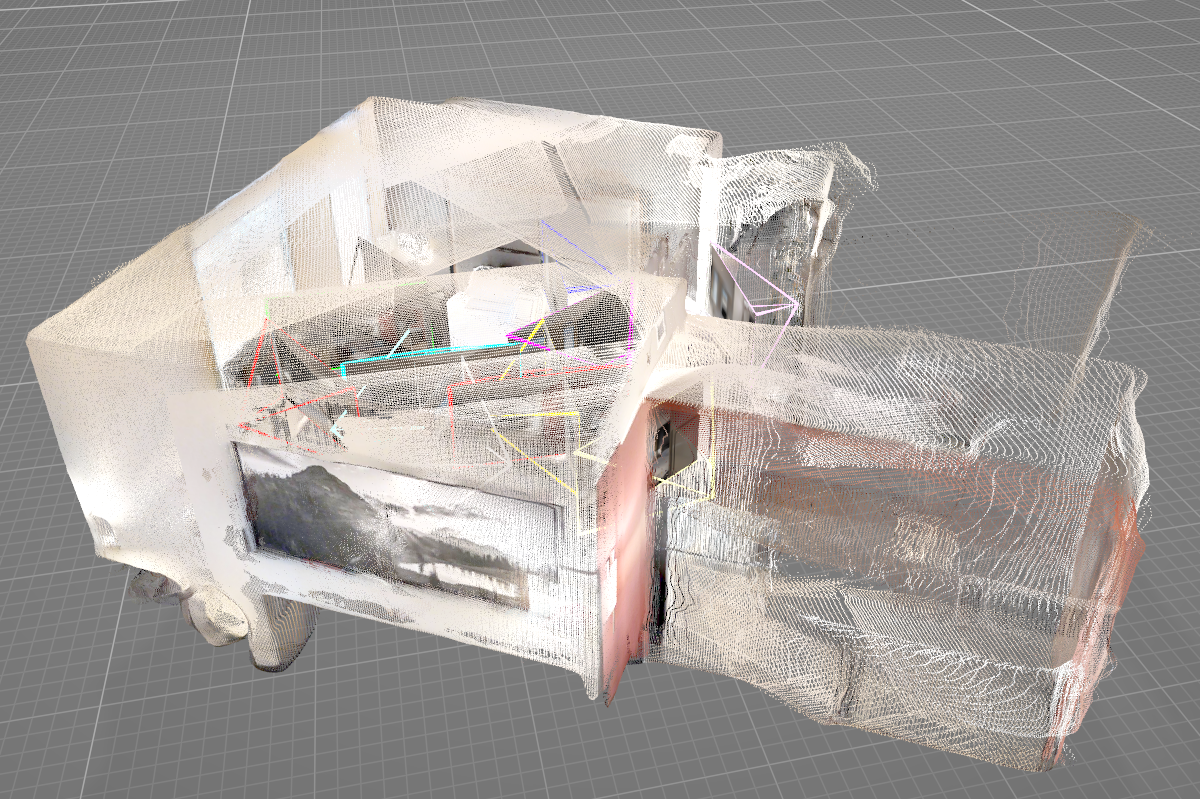}
        \caption{Complete - Scene 1}
    \end{subfigure}
    \hspace{5pt}
    \begin{subfigure}[b]{0.22\textwidth}
        \centering
        \includegraphics[width=\textwidth]{figs/gt_scene6.png}
        \caption{GT - Scene 1}
    \end{subfigure}

    \begin{subfigure}[b]{0.22\textwidth}
        \centering
        \includegraphics[width=\textwidth]{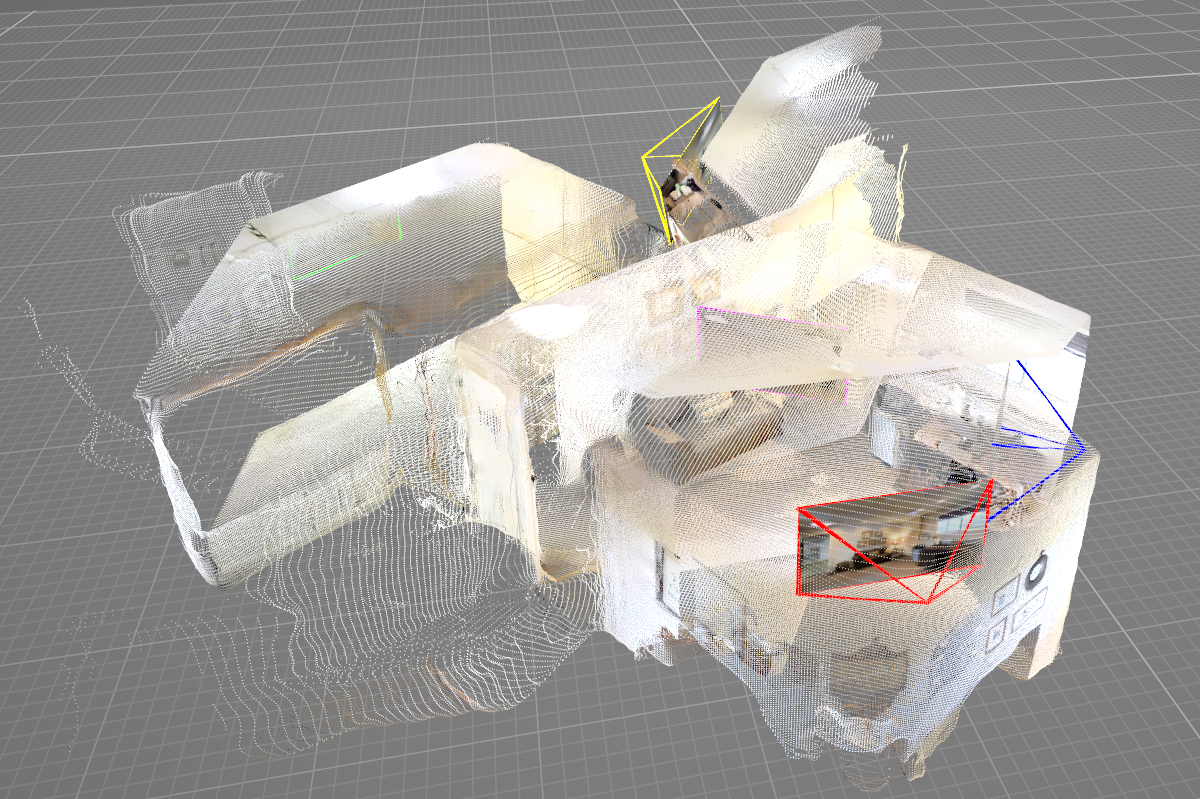}
        \caption{Star - Scene 2}
    \end{subfigure}
    \hspace{5pt}
    \begin{subfigure}[b]{0.22\textwidth}
        \centering
        \includegraphics[width=\textwidth]{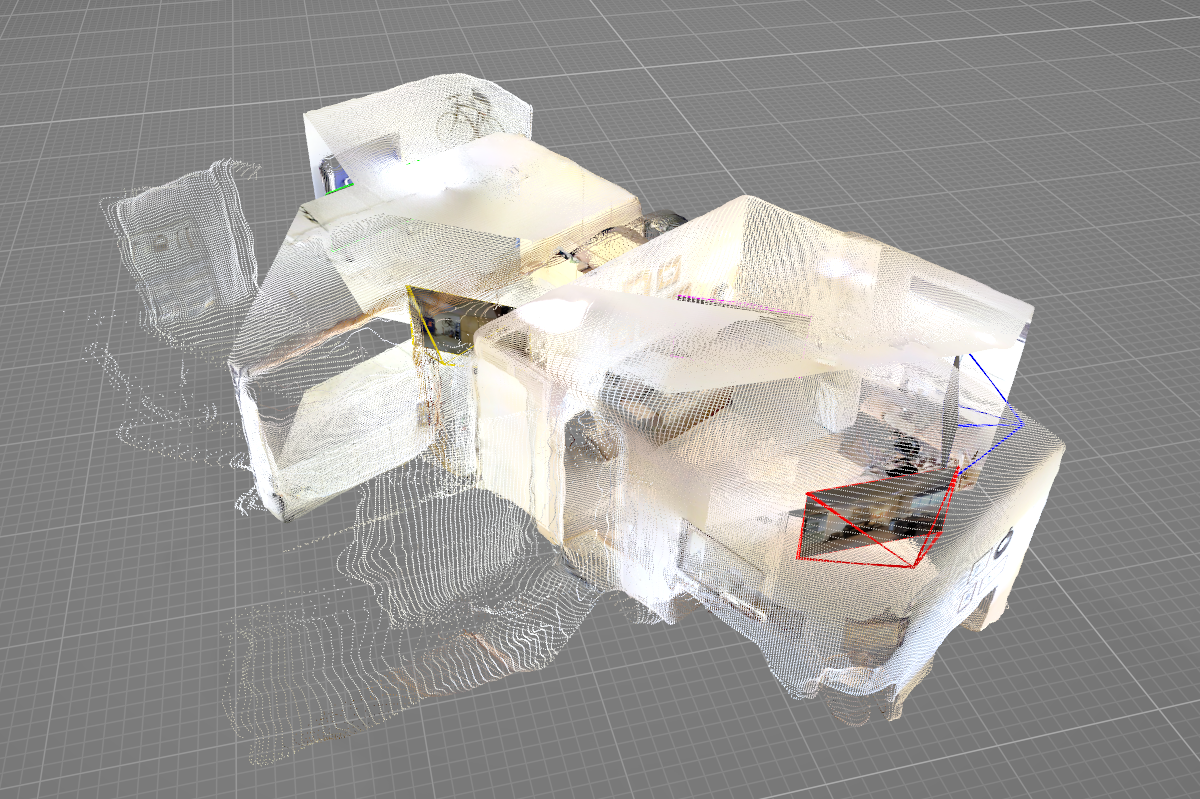}
        \caption{Covision - Scene 2}
    \end{subfigure}
    \hspace{5pt}
    \begin{subfigure}[b]{0.22\textwidth}
        \centering
        \includegraphics[width=\textwidth]{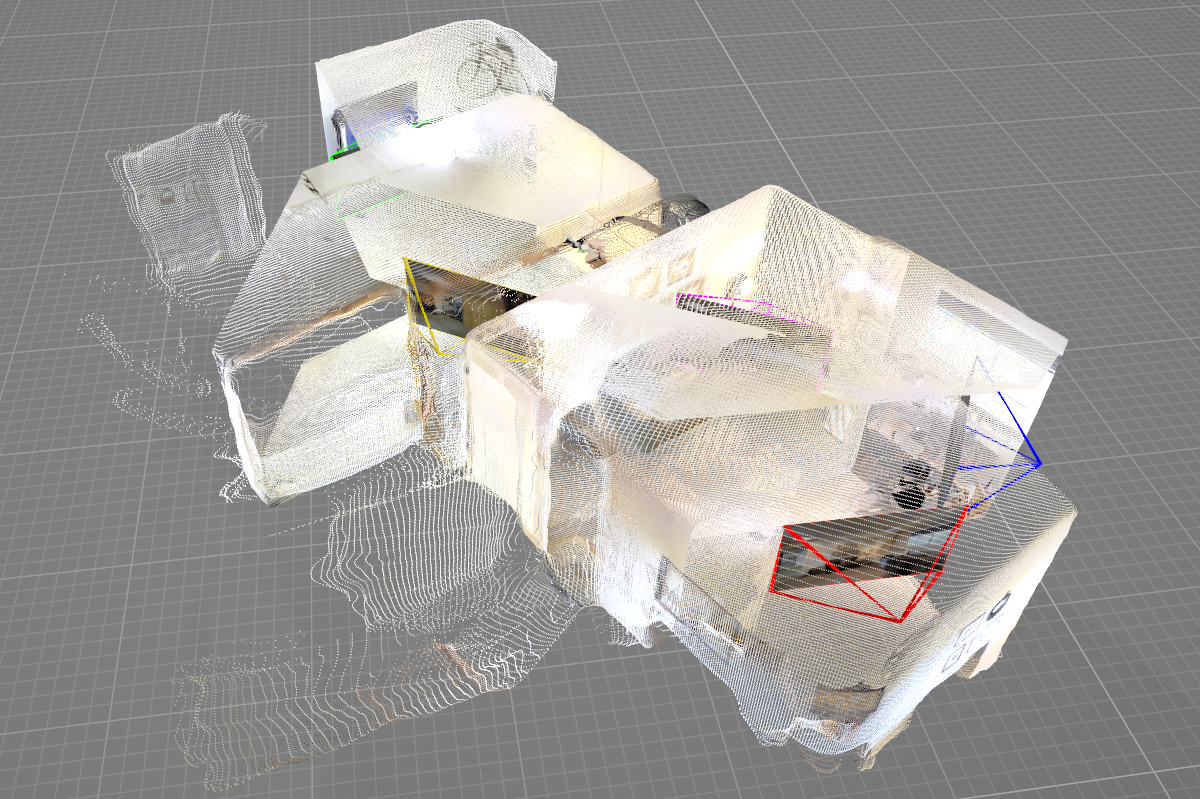}
        \caption{Complete - Scene 2}
    \end{subfigure}
    \hspace{5pt}
    \begin{subfigure}[b]{0.22\textwidth}
        \centering
        \includegraphics[width=\textwidth]{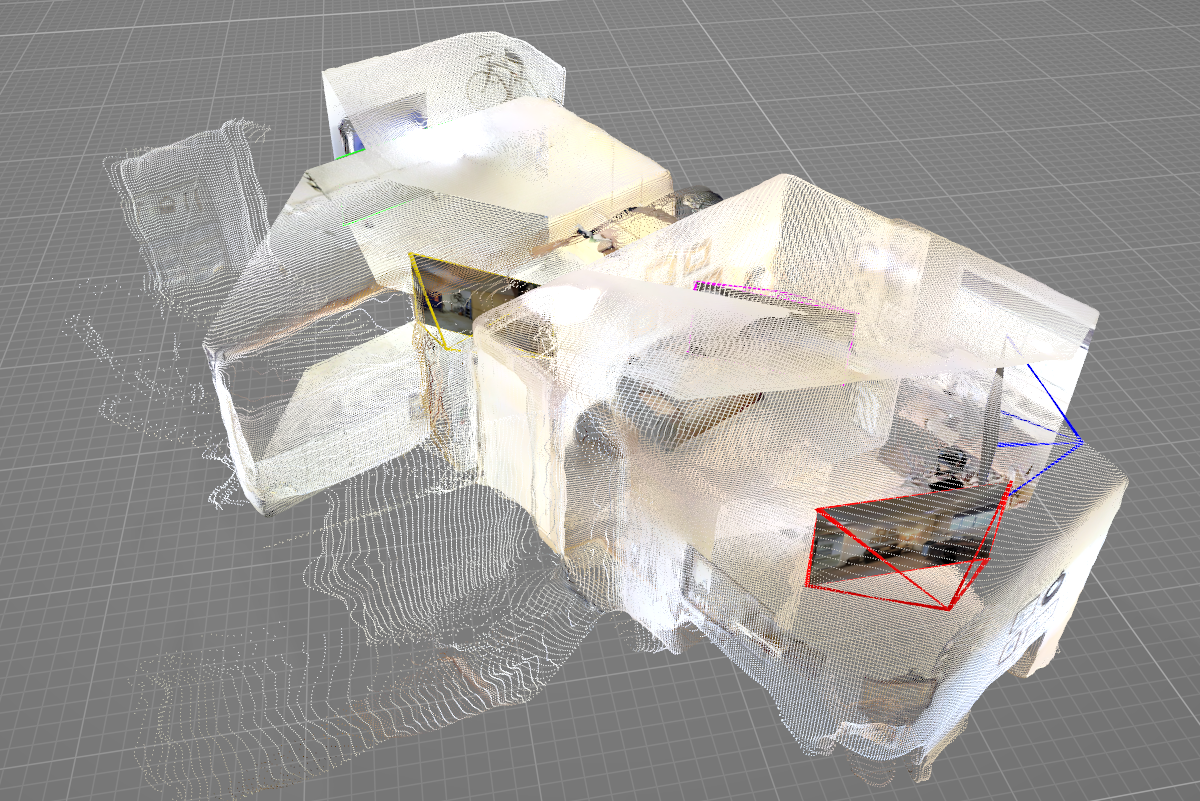}
        \caption{GT - Scene 2}
    \end{subfigure}

    \caption{Two examples of 3D reconstruction results by DUSt3R, each based on distinct graphs.}

    \label{fig:image_pairing_methods}
\end{figure*}

\noindent In addition, we present a comparison of these graphs' performance in terms of 3D reconstruction accuracy and computational cost. As shown in \cref{fig:image_pairing_methods}, we visualize how each graph affects image pairing during reconstruction. The advantages and limitations of each graph type, including aspects such as memory usage and scalability, are discussed in detail in~\cref{sec:dust3r}. This analysis helps justify our choice of the co-visibility graph as the optimal solution for balancing accuracy and computational efficiency.

\section{Detailed definition of different graphs used in CroCo experiment} We train a model to evaluate its performance over the co-visible regions defined by different graph structures, including: (1) High Co-visibility Graph, (2) Co-visibility Graph, and (3) Random Graph.

\begin{itemize}

    \item \textbf{High co-visibility graph}: This graph is created using the pairs of images from a scenario with high spatial and visual overlap ($\ge$50\%, or IoU \textit{i}). Using a threshold \textit{i}, it selects pairs with significant overlap, focusing on high-confidence spatial relationships. This graph aims to test the CroCo model's capability to reconstruct a masked target image using a reference image from a different perspective, particularly emphasizing close spatial relationships.

    \item \textbf{Co-visibility graph}: The Co-Visibility graph of a scenario is the same graph as discussed in \cref{sec:definition} and \cref{sec:dataset}. This usually contains image pairs with both high and low visual/spatial overlap connected by an edge. From these pairs, we randomly select the same number of edges as in the High Co-visibility Graph. This selection ensures a balanced comparison, containing a mixture of high and low co-visibility pairs. This graph is useful in testing the CroCo model's ability to reconstruct images from a variety of perspectives, leveraging both high and low co-visibility information.

    \item \textbf{Random graph}: The Random graph is constructed by selecting all possible image pairs within a scenario, including those with little to no spatial or visual overlap. A subset of these pairs is randomly selected, ensuring the number of edges matches that of the High Co-visibility Graph. This diverse selection, which includes pairs with minimal or no overlap, tests the CroCo model's ability to handle a variety of image combinations, challenging it to perform well across more unpredictable scenarios.

\end{itemize}

\section{Ablation Study of CoVis}\label{sec:abla_covis}

\subsection{Zero-shot Evaluation}
We further evaluate the generalization ability of our Covis by training on one dataset (either Gibson or HM3D) and testing on both Gibson and HM3D individually. As shown in~\cref{tab:cross_dataset_generalization}, Multi-view Covis consistently outperforms pairwise Covis in both in-domain and zero-shot settings, demonstrating better robustness to dataset's domain shifts. We also evaluate the predicted masks using the standard IoU metric (not graph IoU), and achieve an average IoU of 67.3\% against the binarized ground-truth masks.

\begin{table}
\caption{Zero-shot evaluation of multi-view Covis and pairwise Covis across different training and test sets.}
\centering
\vspace{-2mm}
\scriptsize
\resizebox{0.48\textwidth}{!}{
\begin{tabular}{ll|cc|cc}
\toprule[1pt]
\multirow{2}{*}{\textbf{Train}} & \multirow{2}{*}{\textbf{Test}} 
& \multicolumn{2}{c|}{\textbf{Multi-view Covis}} 
& \multicolumn{2}{c}{\textbf{Pairwise Covis}} \\\cline{3-6}
& & IoU & AUC & IoU & AUC \\
\hline
Gibson & Gibson & 0.56 & 0.52 & 0.51 & 0.47 \\
Gibson & HM3D   & 0.51 & 0.48 & 0.48 & 0.44 \\
HM3D   & Gibson & 0.58 & 0.53 & 0.55 & 0.50 \\
HM3D   & HM3D   & 0.52 & 0.48 & 0.50 & 0.46 \\
\bottomrule[1pt]
\end{tabular}
}
\vspace{-2mm}
\label{tab:cross_dataset_generalization}
\end{table}

\subsection{Qualitative Result of Mask Prediction}
\begin{figure}[ht]
    \centering

    \begin{subfigure}[ht]{0.23\textwidth}
        \centering
        \includegraphics[width=\textwidth]{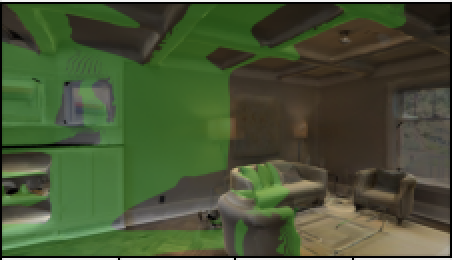}
        \caption{Scene 1 with Green Highlighted Overlap from Scene 2}
    \end{subfigure}
    \begin{subfigure}[ht]{0.23\textwidth}
        \centering
        \includegraphics[width=\textwidth]{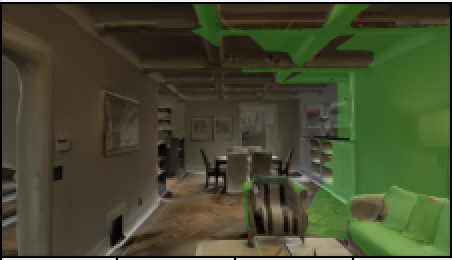}
        \caption{Scene 2 with Green Highlighted Overlap from Scene 1}
    \end{subfigure}

    \begin{subfigure}[ht]{0.23\textwidth}
        \centering        \includegraphics[width=1.01\textwidth,height=2cm]{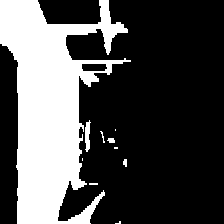}
        \caption{Binarized GT mask for (a)}
    \end{subfigure}
    \begin{subfigure}[ht]{0.23\textwidth}
        \centering
        \includegraphics[width=1.01\textwidth,height=2cm]{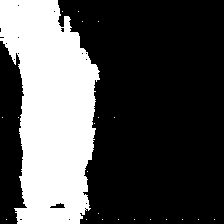}
        \caption{Binary predicted mask for (a)}
    \end{subfigure}
    \vspace{-1mm}
    \caption{Visualization of the co-visible region between a pair of images (a) and (b), with (c) showing the binarized ground truth mask and (d) the predicted mask on the image from (a).}

    \label{fig:covisibility tasks}
\end{figure}

We also visualize the ground-truth co-visibility masks between image pairs; which serve as additional supervision signals during training, and the binary co-visibility masks learned by Covis model in~\cref{fig:covisibility tasks}.

\section{Sim2Real} 
\begin{figure}[!h]
    \vspace{2mm}  
    \centering
    \begin{subfigure}[b]{0.5\textwidth}\centering\includegraphics[width=\textwidth]{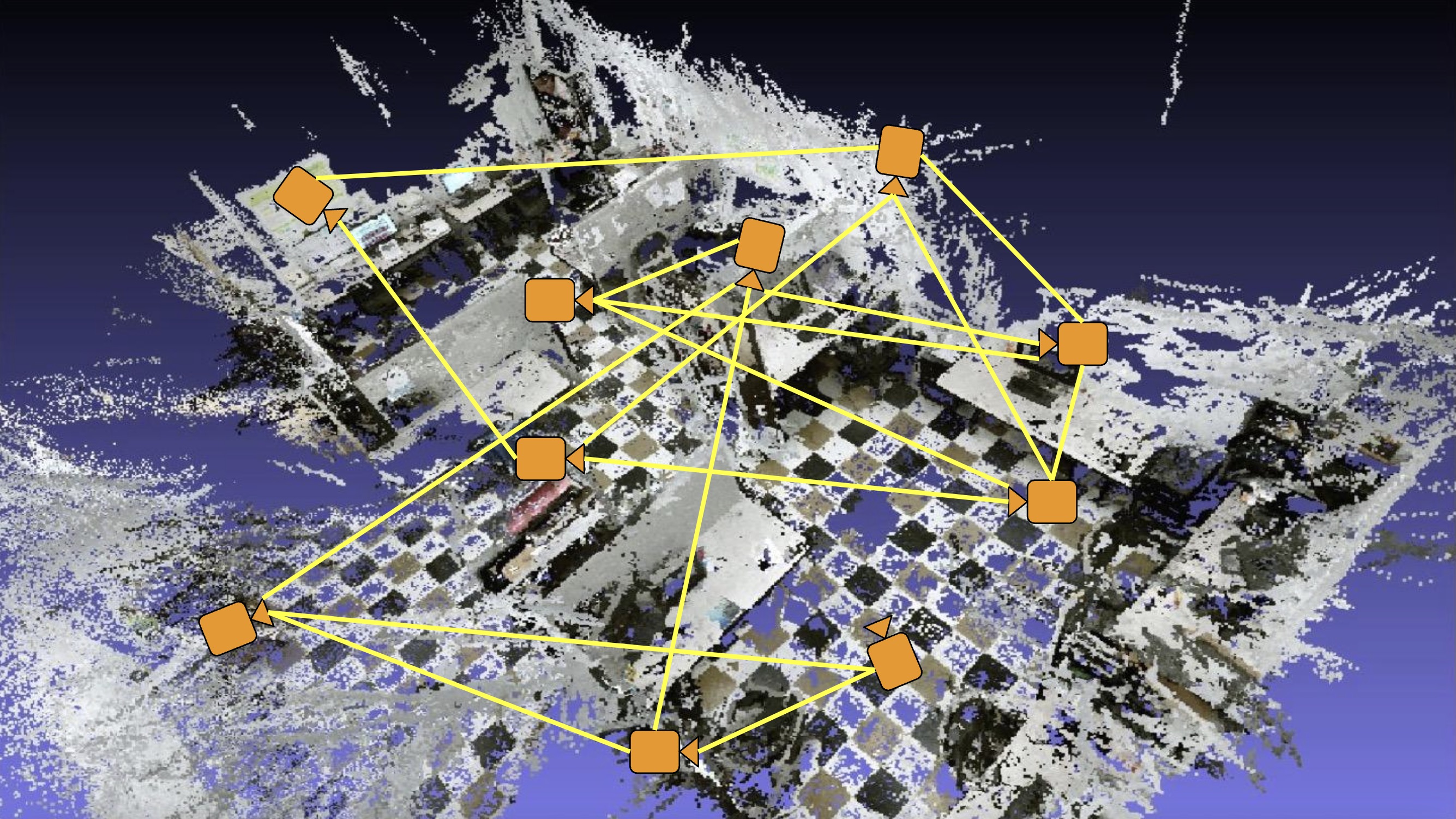}
        \caption{Manually Labeled Topology}
    \end{subfigure}
    \hspace{10pt} 
    \begin{subfigure}[b]{0.5\textwidth}
        \centering\includegraphics[width=\textwidth]{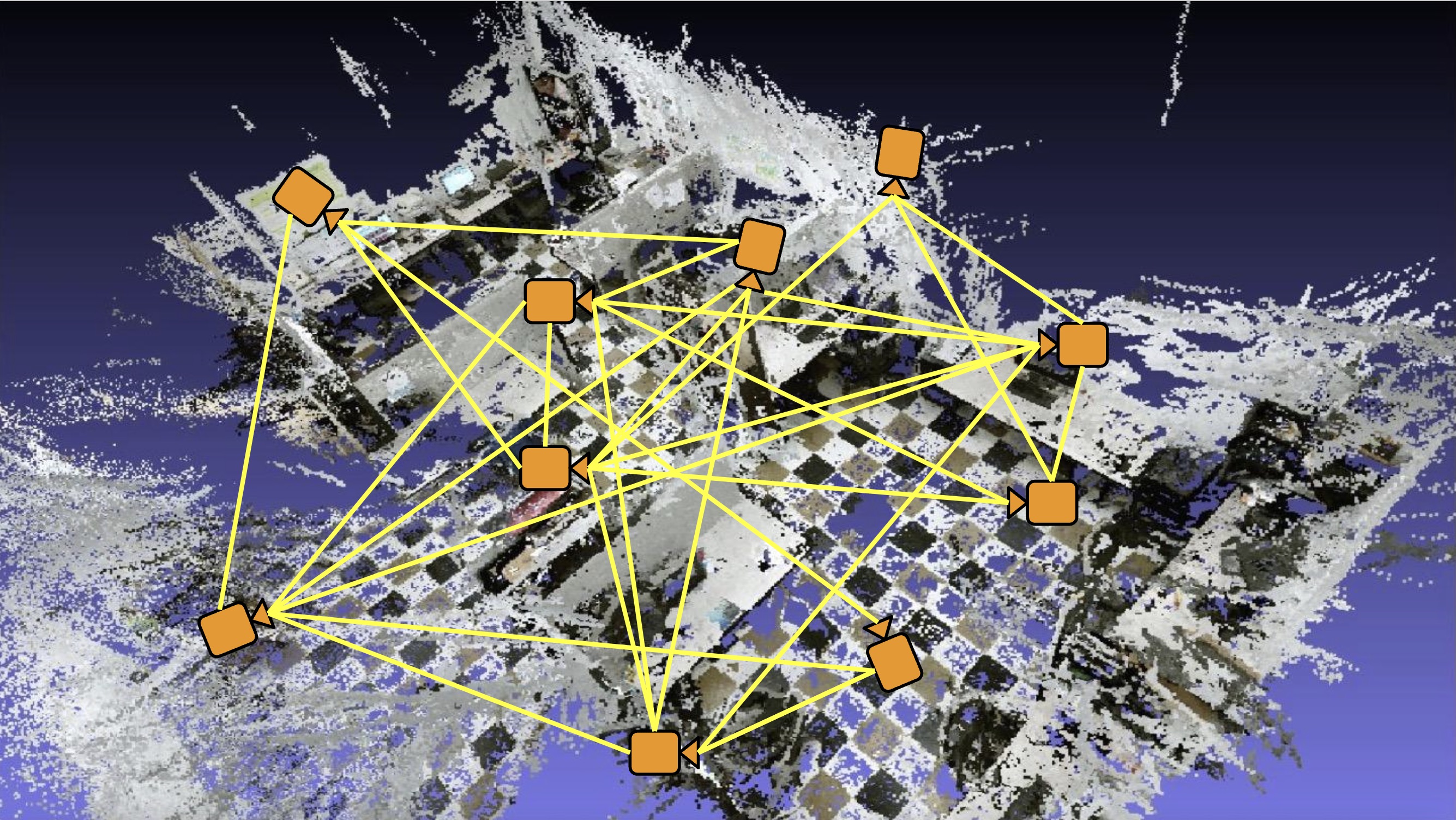}
        \caption{Covis Predicted Topology}
    \end{subfigure}
    \caption{Comparison of manually labeled and Covis predicted Topologies. For clarity, we show 10 sampled images to provide an illustrative example.}
    \label{fig:topology_comparison}
    \vspace{-2mm}
\end{figure}

In this experiment, our aim is to demonstrate that the Co-VisiON dataset exhibits a small domain gap compared to real-world environments, such as the AVD dataset~\cite{ammirato2017dataset}. We have tested Covis, ViT, VGG, Resnet, Contrastive, and NetVlad methods on the AVD dataset using pretrained model from Co-VisiON dataset. We observe AUC metric results in~\cref{table:real_world_auc_results} to be comparable to those in~\cref{tab:table1}. Similarly, from~\cref{fig:topology_comparison}, we can see that the co-visibility graph predicted using the Covis closely resembles the manually labelled topology graph.

\begin{table}[ht]
\vspace{2mm} 
\caption{Zero-shot comparison of AUC values (\%) for baseline models pretrained on the Co-VisiON dataset and evaluated on real-world data.}

\small
\setlength{\tabcolsep}{5pt}
\resizebox{0.48\textwidth}{!}{
\begin{tabular}{lcccccc} \toprule[2pt]
\textbf{Baseline} & \textbf{Covis} &\textbf{ViT} & \textbf{VGG} & \textbf{ResNet} & \textbf{Contrastive} & \textbf{NetVlad} \\ \hline
AUC & 0.61 & 0.52 & 0.33 & 0.32 & 0.31 & 0.22 \\
\bottomrule[2pt] 
\end{tabular}
}
\vspace{-1mm}
\label{table:real_world_auc_results}
\end{table}